UK-RAS NETWORK
ROBOTICS & AUTONOMOUS SYSTEMS

# Agricultural Robotics:
# **The Future of Robotic Agriculture**

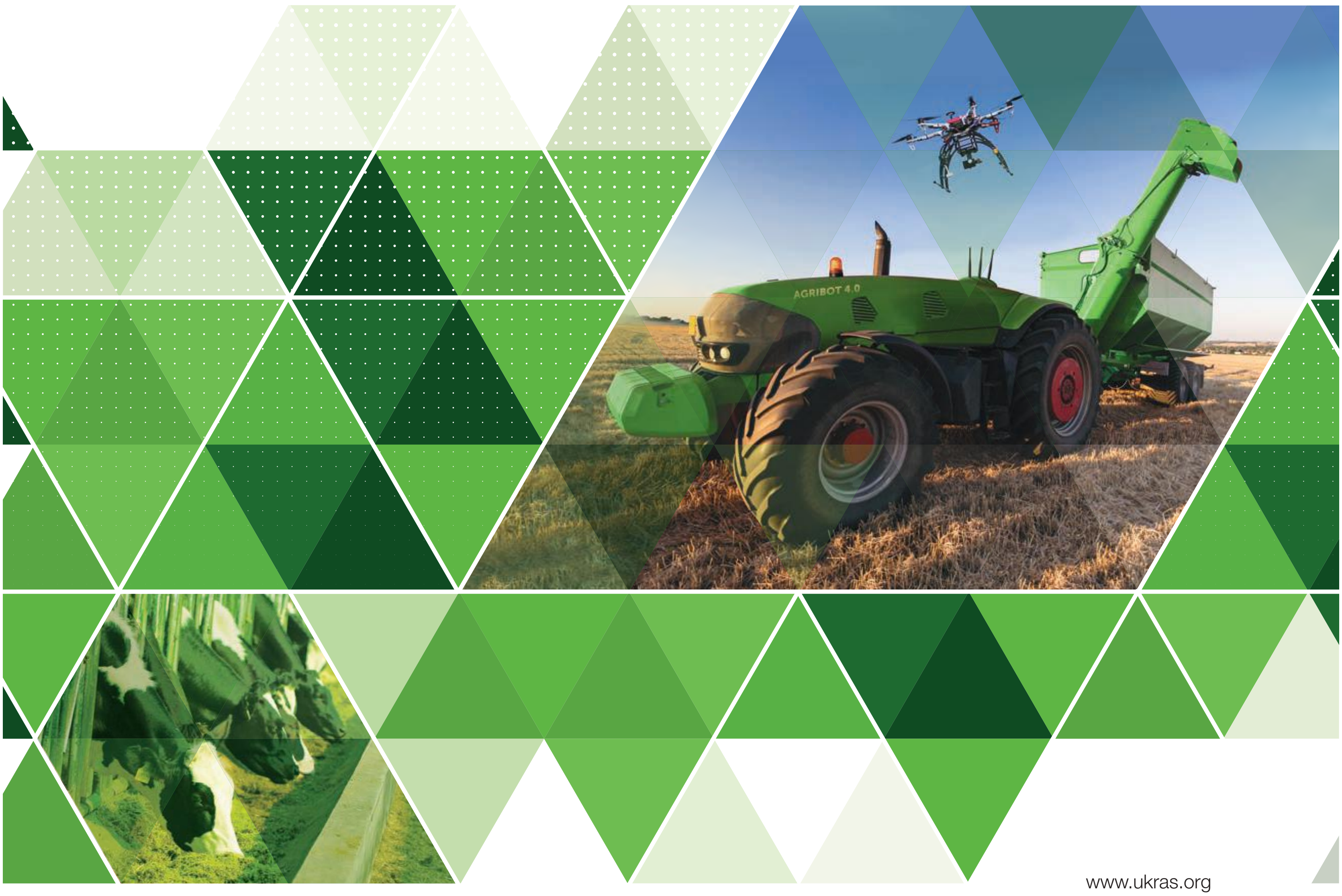

www.ukras.org

UK-RAS NETWORK
ROBOTICS & AUTONOMOUS SYSTEMS

UKRAS.ORG

# FOREWORD

Welcome to the UK-RAS White Paper Series on Robotics and Autonomous Systems (RAS). This is one of the core activities of UK-RAS Network, funded by the Engineering and Physical Sciences Research Council (EPSRC).

By bringing together academic centres of excellence, industry, government, funding bodies and charities, the Network provides academic leadership, expands collaboration with industry while integrating and coordinating activities at EPSRC funded RAS capital facilities, Centres for Doctoral Training and partner universities.

The recent commitment of a £90million investment by the government (Transforming Food Production Challenge through the Industrial Strategy) supports the idea that Agri-tech is a burgeoning market, and we are proud to be exploring the use of robotics in this important sector, employing almost 4 million people and larger than the automotive and aerospace sectors combined. Agri-tech companies are already working closely with UK farmers, using technology, particularly robotics and AI, to help create new technologies and herald new innovations. This is a truly exciting time for the industry as there is a growing recognition that the significant challenges facing global agriculture represent unique opportunities for innovation, investment and commercial growth.

This white paper aims to provide an overview of the current impact and challenges facing Agri-tech, as well as associated ethical considerations. We hope the paper will provide the reader with an overview of the current trends, technological advances, as well as barriers that may impede the sector's full potential. We have included recommendations to some of the challenges identified and hope this paper provides a basis for discussing the future technological roadmaps, engaging the wider community and stakeholders, as well as policy makers, in assessing the potential social, economic and ethical/legal impact of RAS in agriculture.

It is our plan to provide annual updates for these white papers so your feedback is essential - whether it is to point out inadvertent omissions of specific areas of development that need to covered, or to suggest major future trends that deserve further debate and in-depth analysis. Please direct all your feedback to whitepaper@ukras.org. We look forward to hearing from you!

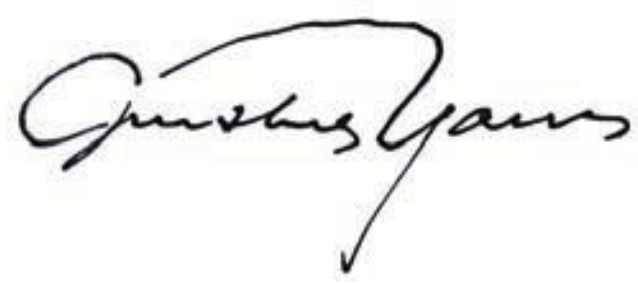

Prof Guang-Zhong Yang, CBE, FREng
Chairman, UK-RAS Network

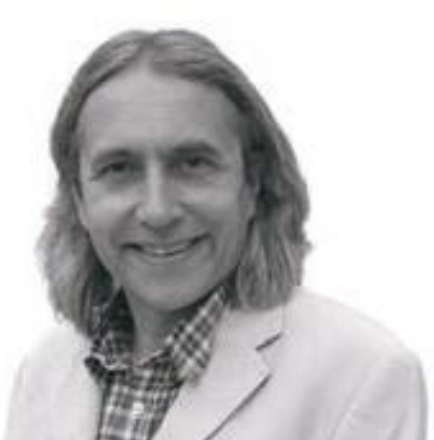

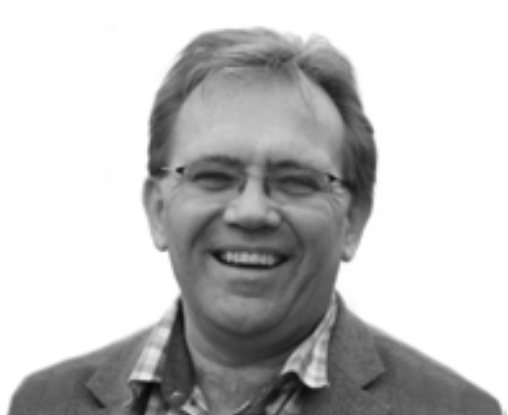

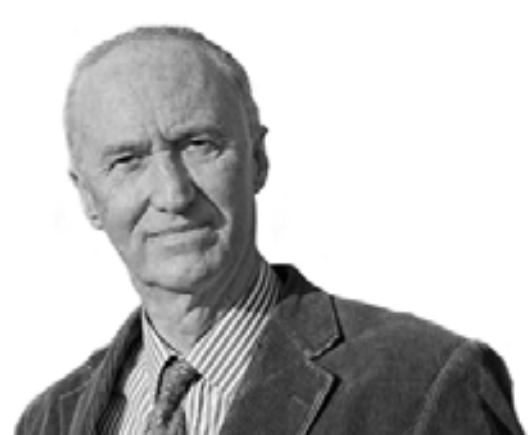

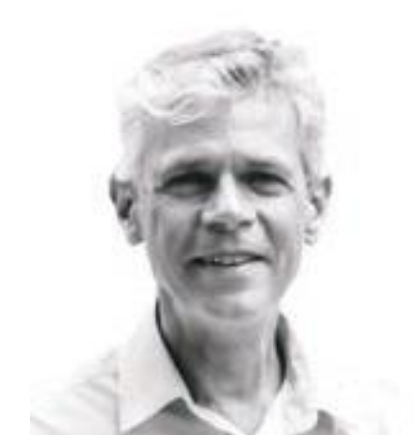

**Tom Duckett,**
University of Lincoln,
tduckett@lincoln.ac.uk

**Simon Pearson,**
University of Lincoln,
spearson@lincoln.ac.uk

**Simon Blackmore**
Harper Adams University,
simon.blackmore@harper-adams.ac.uk

**Bruce Grieve,**
University of Manchester,
Bruce.Grieve@manchester.ac.uk

Contributions: Wen-Hua Chen (Loughborough University), Grzegorz Cielniak (University of Lincoln), Jason Cleaversmith (Akva Group Scotland), Jian Dai (King's College London), Steve Davis (University of Salford), Charles Fox (University of Lincoln), Pål From (University of Lincoln), Ioannis Georgilas (University of Bath), Richie Gill (University of Bath), Iain Gould (University of Lincoln), Marc Hanheide (University of Lincoln), Alan Hunter (University of Bath), Fumiya Iida (Cambridge University), Lyudmila Mihalyova (Sheffield University), Samia Nefti-Meziani (University of Salford), Gerhard Neumann (University of Lincoln), Paolo Paoletti (University of Liverpool), Tony Pridmore (University of Nottingham), Dave Ross (Scotland's Rural College), Melvyn Smith (University of the West of England), Martin Stoelen (University of Plymouth), Mark Swainson (University of Lincoln), Sam Wane (Harper Adams University), Peter Wilson (University of Bath), Isobel Wright (University of Lincoln).

# EXECUTIVE SUMMARY

Agri-Food is the largest manufacturing sector in the UK. It supports a food chain that generates over £108bn p.a., with 3.9m employees in a truly international industry and exports £20bn of UK manufactured goods. However, the global food chain is under pressure from population growth, climate change, political pressures affecting migration, population drift from rural to urban regions and the demographics of an aging global population. These challenges are recognised in the UK Industrial Strategy white paper and backed by significant investment via a Wave 2 Industrial Challenge Fund Investment ("Transforming Food Production: from Farm to Fork"). Robotics and Autonomous Systems (RAS) and associated digital technologies are now seen as enablers of this critical food chain transformation. To meet these challenges, this white paper reviews the state of the art in the application of RAS in Agri-Food production and explores research and innovation needs to ensure these technologies reach their full potential and deliver the necessary impacts in the Agri-Food sector.

The opportunities for RAS range include; the development of field robots that can assist workers by carrying payloads and conduct agricultural operations such as crop and animal sensing, weeding and drilling; integration of autonomous systems technologies into existing farm operational equipment such as tractors; robotic systems to harvest crops and conduct complex dextrous operations; the use of collaborative and "human in the loop" robotic applications to augment worker productivity; advanced robotic applications, including the use of soft robotics, to drive productivity beyond the farm gate into the factory and retail environment; and increasing the levels of automation and reducing the reliance on human labour and skill sets, for example, in farming management, planning and decision making.

RAS technology has the potential to transform food production and the UK has an opportunity to establish global leadership within the domain. However, there are particular barriers to overcome to secure this vision:

1. The UK RAS community with an interest in Agri-Food is small and highly dispersed. There is an urgent need to defragment and then expand the community.
2. The UK RAS community has no specific training paths or Centres for Doctoral Training to provide trained human resource capacity within Agri-Food.
3. While there has been substantial government investment in translational activities at high Technology Readiness Levels (TRLs), there is insufficient ongoing basic research in Agri-Food RAS at low TRLs to underpin onward innovation delivery for industry.
4. There is a concern that RAS for Agri-Food is not realising its full potential, as the projects being commissioned currently are too few and too small-scale. RAS challenges often involve the complex integration of multiple discrete technologies (e.g. navigation, safe operation, grasping and manipulation, perception). There is a need to further develop these discrete technologies but also to deliver large-scale industrial applications that resolve integration and interoperability issues. The UK community needs to undertake a few well-chosen large-scale and collaborative "moon shot" projects.
5. The successful delivery of RAS projects within Agri-Food requires close collaboration between the RAS community and with academic and industry practitioners. For example, the breeding of crops with novel phenotypes, such as fruits which are easy to see and pick by robots, may simplify and accelerate the application of RAS technologies. Therefore, there is an urgent need to seek new ways to create RAS and Agri-Food domain networks that can work collaboratively to address key challenges. This is especially important for Agri-Food since success in the sector requires highly complex cross-disciplinary activity. Furthermore, within UKRI many of the Research Councils and Innovate UK directly fund different aspects of Agri-Food, but as yet there is no coordinated and integrated Agri-Food research policy per se.

Our vision is a new generation of smart, flexible, robust, compliant, interconnected robotic and autonomous systems working seamlessly alongside their human co-workers in farms and food factories. Teams of multi-modal, interoperable robotic systems will self-organise and coordinate their activities with the "human in the loop". Electric farm and factory robots with interchangeable tools, including low-tillage solutions, soft robotic grasping technologies and sensors, will support the sustainable intensification of agriculture, drive manufacturing productivity and underpin future food security.

To deliver this vision the research and innovation needs include the development of robust robotic platforms, suited to agricultural environments, and improved capabilities for sensing and perception, planning and coordination, manipulation and grasping, learning and adaptation, interoperability between robots and existing machinery, and human-robot collaboration, including the key issues of safety and user acceptance.

Technology adoption is likely to occur in measured steps. Most farmers and food producers will need technologies that can be introduced gradually, alongside and within their existing production systems. Thus, for the foreseeable future, humans and robots will frequently operate collaboratively to perform tasks, and that collaboration must be safe. There will be a transition period in which humans and robots work together as first simple and then more complex parts of work are conducted by robots, driving productivity and enabling human jobs to move up the value chain.

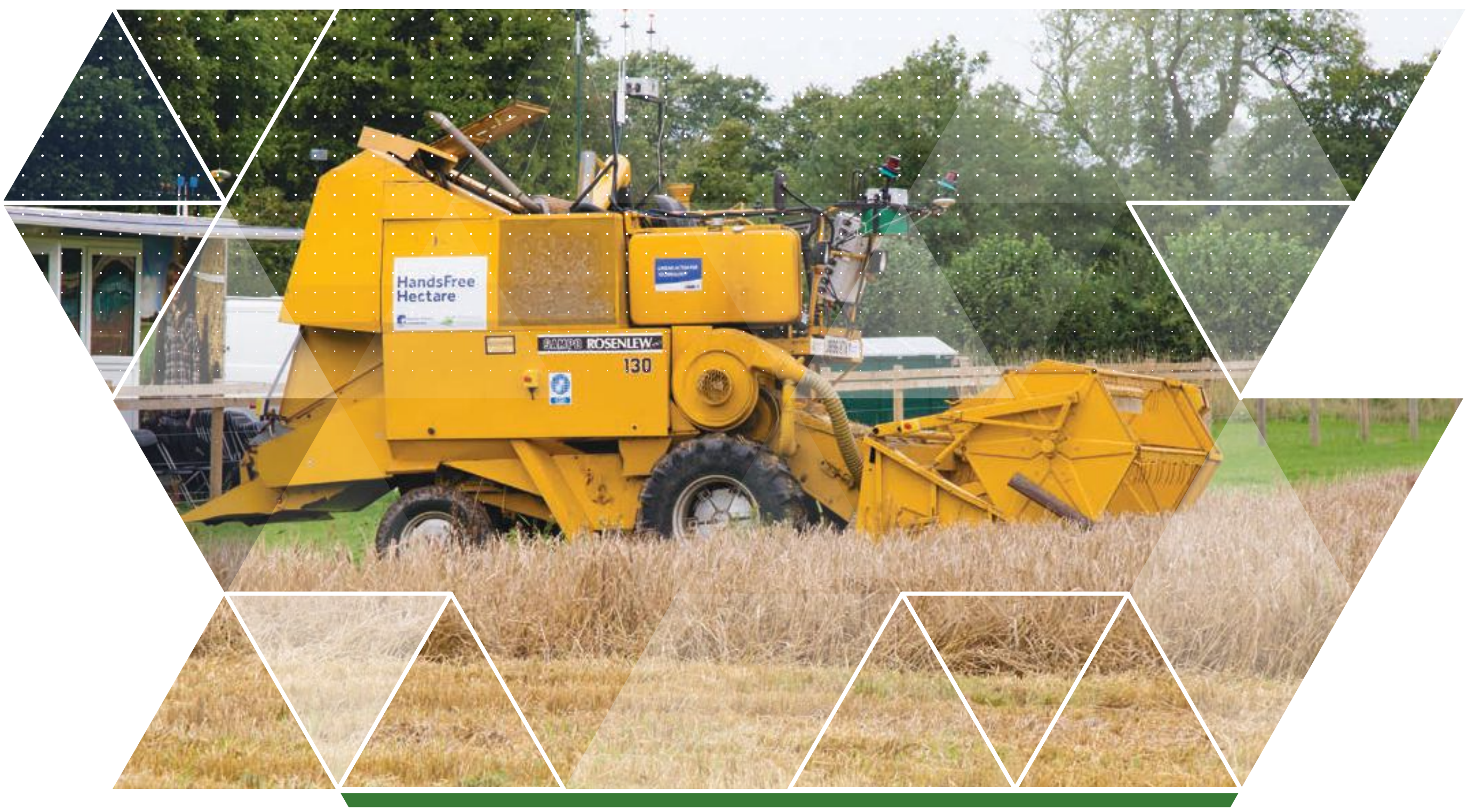

"Robotics and Autonomous Systems (RAS) are set to transform global industries. These technologies will have greatest impact on large sectors of the economy with relatively low productivity such as Agri-Food (food production from the farm through to and including manufacturing to the retail shelf). The UK Agri-Food chain, from primary farming through to retail, generates over £108bn p.a., with 3.7m employees in a truly international industry yielding £20bn of exports in 2016."

# CONTENTS

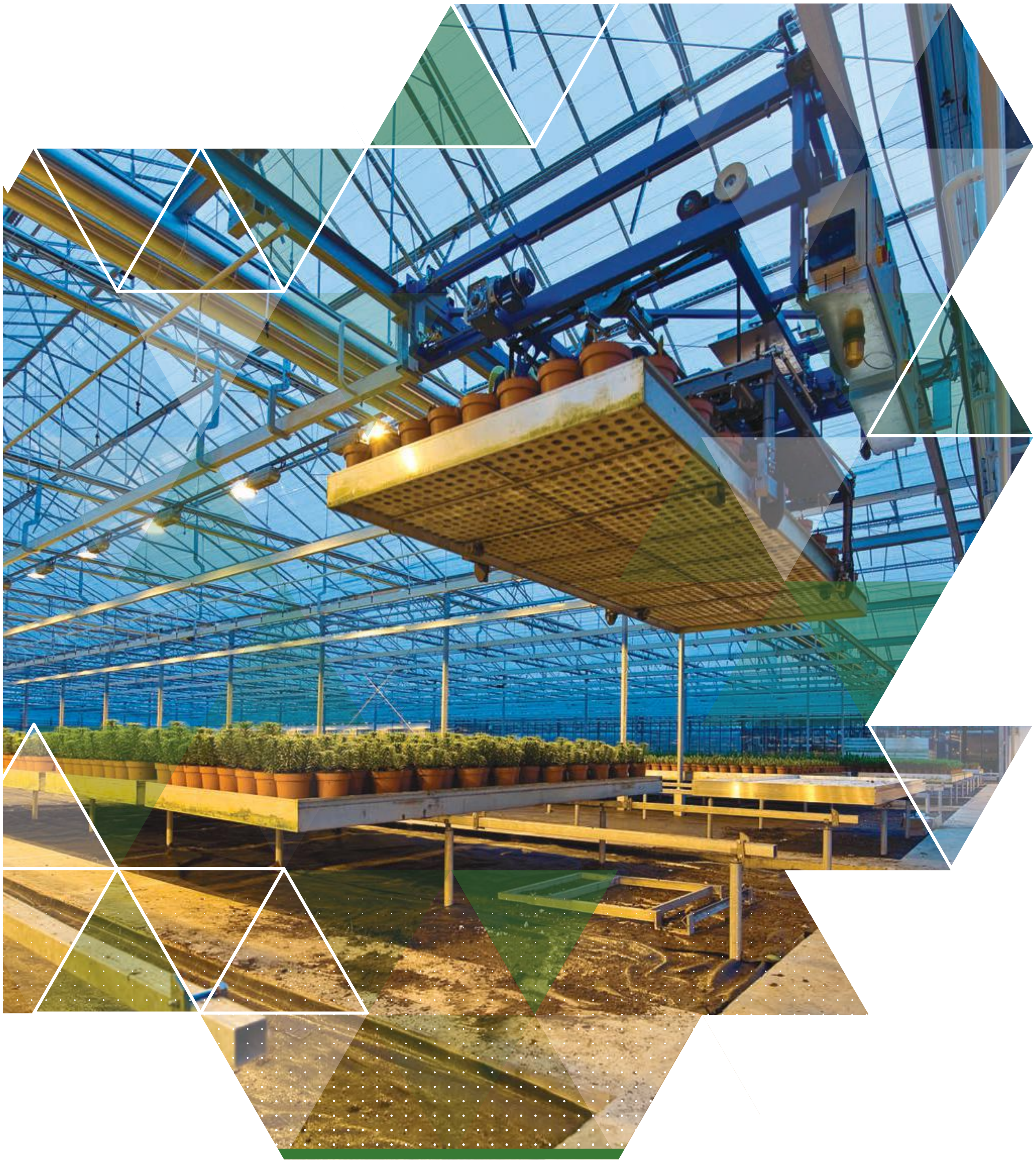

# 1. ECONOMIC, SOCIAL AND ENVIRONMENTAL DRIVERS

## 1.1 ECONOMIC AND SOCIETAL FACTORS

Robotics and Autonomous Systems (RAS) are set to transform many global industries. These technologies will have greatest impact on large sectors of the economy with relatively low productivity such as Agri-Food (food production from the farm through to and including manufacturing to the retail shelf). The UK Agri-Food chain, from primary farming through to retail, generates over £108bn p.a., with 3.7m employees in a truly international industry yielding £20bn of exports in 2016 [1].

The global food chain cannot be taken for granted: it is under pressure from global population growth and needs to drive productivity, climate change, inescapable political impacts of migration (e.g. Brexit and potential US migration restrictions), population drift from rural to urban regions, and the demographics of an aging global population in advanced economies including China. In the UK the uncertainty associated with Brexit is already affecting migrant worker confidence and availability. These issues, manifest via different mechanisms (demographics, urban drift, etc.), are now impacting on many sectors of the global Agri-Food industry. In addition, jobs in the Agri-Food sector can be physically demanding, repetitive in nature, conducted in adverse environments and relatively unrewarding.

Given these circumstances the global Agri-Food sector could be transformed by advanced RAS technologies. The recent Made Smarter Review [2] considered that digital technologies, including RAS, deployed in food manufacturing alone could add £58 bn of GVA to the UK economy over the next 13 years. Robotic automation would also help to attract skilled workers and graduates to the sector. These opportunities have been further recognised by national government and are outlined in the Industrial Strategy White Paper [3]. The Secretary of State for BEIS announced in late February 2018 a £90m ISCF Wave 2 investment (Transforming food production: from farm to fork) to support innovation, including robotics and digital systems to drive innovation in the Agri-Food chain. This follows an earlier £160m investment that funded the UK Agri-Tech Catalyst program (now near completion).

## 1.2 ENVIRONMENTAL BENEFITS

As well as delivering economic benefits, such as increasing productivity and reducing waste throughout the food supply chain, developing a new focus for RAS within Agri-Food will have significant societal and environmental benefits. For example, the food chain uses 18% of UK energy consumption [4], while high mass farm machinery is causing unsustainable compaction damage to our soils; meanwhile, many species of British wildlife face risk of extinction due to modern farming practices [5], including widespread use of herbicides and pesticides, industrialisation of machinery, and reduction in hedgerows and drainage due to increasing field sizes.

### Soils

There is a wide variety of soil types across the UK, exhibiting a range of properties to consider (e.g. texture, pH, fertility status). The profile of the soil in many parts of the UK is defined by clay, gravel or weathered bedrock horizons with a relatively shallow fertile topsoil above, making it imperative that agricultural machinery does not cause damage such as compaction or erosion. Compaction can lead to reductions in crop yields, and a need for greater fertiliser and fuel inputs. The total losses from degradation to soils in England and Wales alone have been estimated at around £1.2bn p.a. [6]. Furthermore, soil compaction has wider environmental costs; increasing waterlogging, surface run-off and nitrous oxide emissions, and restricting the habitat for soil fauna [7]. Fleets of small lightweight robots are now seen as a replacement for traditional high mass tractors, allowing a gradual reduction of compaction, re-aeration of the soil and benefits to soil function.

### Water

Agriculture uses 70% of all global fresh water supplies, and yet 4bn people live in global regions with water scarcity [8]. In the UK, weather can play an unpredictable role in agriculture with short periods of drought or flooding in many rural areas. This unpredictability makes it difficult to put in fixed systems of drainage or irrigation at a justifiable cost. The issues of water are complex: we need to find new means to drive water use efficiency within agricultural systems, while onward diffuse pollution to water bodies has serious negative environmental impacts. DEFRA estimates that the UK cost of diffuse pollution (nitrate, phosphorous, pesticide and sediment run-off, etc.) amounts to c. £311m per annum [9]. Field robots are already being deployed to help farmers measure, map and optimise water and irrigation use. Likewise robots that use precision technologies to apply fertilisers and pesticides within agricultural systems will reduce environmental impacts.

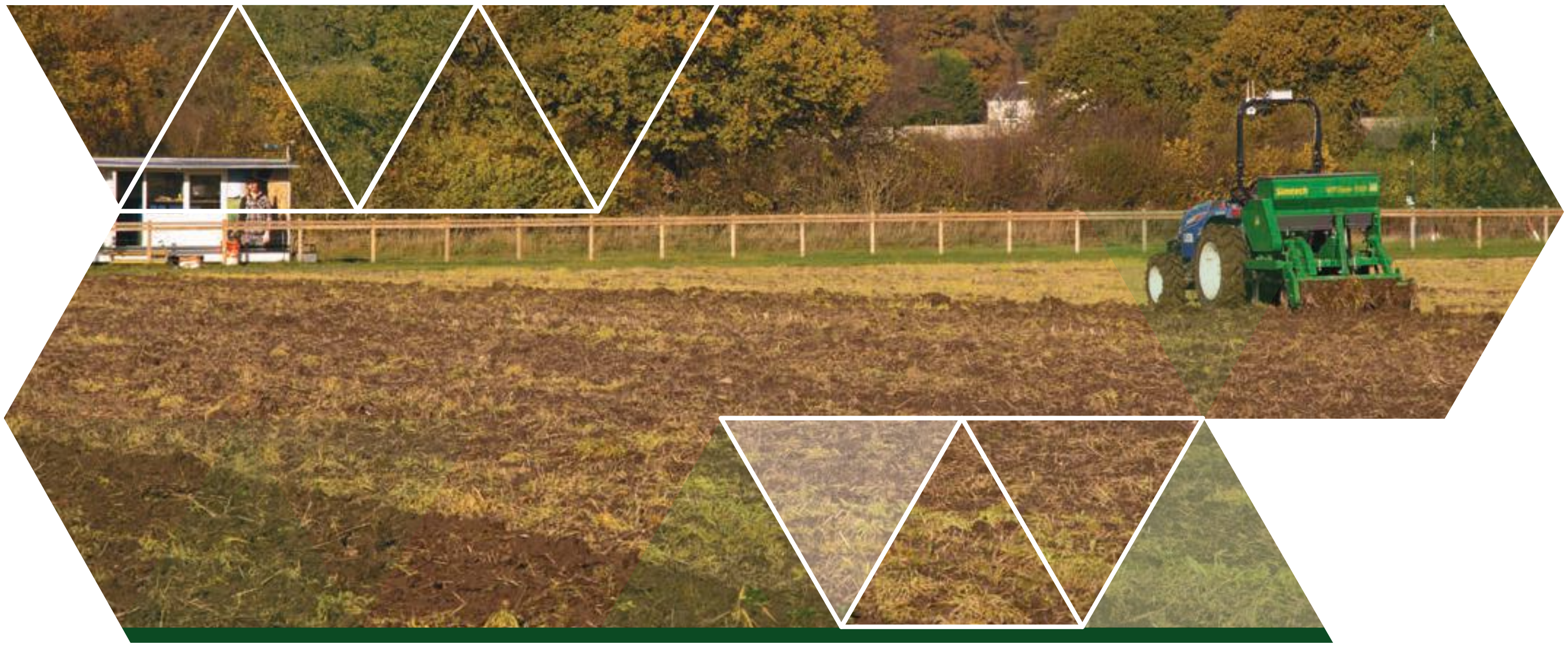

**Pesticides**
Agricultural systems globally are now highly reliant on the large-scale application of synthetic pesticides to control weeds, insects and diseases. An improved understanding of chemical safety and environmental impacts has led to multiple product withdrawals, reducing the number of active ingredients available to farmers. Furthermore, ever higher regulation and registration costs has reduced the number of new pesticides entering the agricultural market. There is therefore a global need to find new ways to produce crops that do not require or reduce the use of pesticides. There are now a number of crop weeding robots that reduce the need for herbicides by deploying camera-guided hoes [10], precision sprayers [11] or lasers [12] to manage weeds. Although in its infancy, this technology shows great promise. In addition, novel sensors deployed on robots can reduce pesticide use by both detecting pests and diseases and precisely targeting the application of insecticides and fungicides. Robots could also be deployed as part of integrated pest management systems, for example, for the accurate and low-cost dispersal of biopesticides to counteract crop pests and diseases.

**Electrification of farm vehicles and implements**
The proliferation of electric motors and actuators in applications ranging from industrial processes to modern passenger aircraft and cars is indicative of the drive away from mechanical traction and actuation systems to electrically based systems. Large diesel vehicles are likely to remain in practical use for many years to come, however the optimum use of such vehicles has been to go as large as possible, which in itself leads to issues such as soil compaction and “brute force” delivery of fertilisers, herbicides and pesticides. The migration from monolithic, fossil-fuel-based agricultural platforms to fleets of smaller electric powered robotic platforms offers the possibility of much lower emissions with locally generated power. Recent years have seen an increase in the use of agricultural land for solar photovoltaic, wind turbines and anaerobic digestion plants. Therefore, the potential for dual use of not only the land, but also the electricity generated, is of interest to the agricultural robotics community. Many agricultural implements are driven directly from the prime mover (often a tractor) via a mechanical linkage. By using electric drives the efficiency can be much higher and the whole system made safer as a result. One of the most common sources of injury and death on the farm is the mechanical linkages in large farm machinery. Therefore there are potentially major health and safety benefits to electrification and automation of farm equipment.

## 1.3 PRECISION AGRICULTURE

Also known as ‘smart farming’, precision agriculture has its origins in developments first applied in industrial manufacturing as far back as the 1970s and 80s. It concerns the use of monitoring and intervention techniques to improve efficiency, realised in application through the deployment of sensing technologies and automation. The development of precision agriculture has been driven by the desire to better handle the spatial and temporal variability, e.g. in soil water-content or crop varieties, from farm-scale, down to field-scale, through to sub-field scale [13]. One approach is to utilise more intelligent machines to reduce and target inputs in more effective ways. The advent of autonomous system architectures gives us the opportunity to develop a new range of flexible agricultural equipment based on small, smart machines that reduces waste, improves economic viability, reduces environmental impact and increases food sustainability. There is also considerable potential for robotics technologies to increase the window of opportunity for intervention, for example, being able to travel on wet soils, work at night, etc.

Sensory data collected by robotic platforms in the field can further provide a wealth of information about soil, seeds, livestock, crops, costs, farm equipment and the use of water and fertiliser. Low-cost Internet of Things (IoT) technologies and advanced analytics are already beginning to help farmers analyse data on weather, temperature, moisture, prices, etc., and provide insights into how to optimise yield, improve planning, make smarter decisions about the level of resources needed, and determine when and where to distribute those resources in order to minimise waste and increase yields [14]. Future telecommunications availability is likely to enhance IoT capacity, with agri-tech test beds already under development.

## 1.4 LIVESTOCK AND AQUACULTURE

**Robotics and autonomous systems for livestock**
At the farm level, robotic systems are now commonly deployed for milking animals [15]. The take-up is a relatively small percentage at the moment, but an EU foresight study predicts that around 50% of all European herds will be milked by robots by 2025 [16]. Robotic systems are starting to perform tasks around the farm, such as removing waste from animal cubicle pens, carrying and moving feedstuffs, etc. Systems are in use and under development for autonomously monitoring livestock and collecting field data, all commercially useful for efficient and productive livestock farming. There are further opportunities to apply more advanced sensor technologies, combined with more autonomous systems, to perform tasks on the farm. This applies to both extensive production and intensive (indoor) systems. Extensive livestock utilise c. 45% of UK's area that is grassland and not fit for crop production, and derive food products from this resource. Thus management of this feed resource is also important.

A further application for robotic systems concerns the management of farmed animals, such as dairy cattle, pigs and chickens, where intervention via the provision of appropriate and timely data can help reduce waste and environmental pollution as well as improve animal welfare and productivity on the farm. Welfare accreditation schemes and measures undertaken for assurance purposes (RSPCA Assured, Red Tractor, etc.) rely heavily on systems-based approaches by inspectors observing the farm intermittently at a group level. Precision farming on the other hand has the potential to offer animal-centric health and welfare assessment that would operate continuously to assess the condition and state of individual animals. Other wider societal benefits may follow in terms of improved working conditions, a stronger more competitive UK agricultural sector (via improve feed conversion efficiency per kg of meat produced) and better food security, providing consumers with greater access to lower-cost, higher quality meat products.

Farmers must constantly monitor their animals and their setting in order to ensure animal health and maintain a comfortable, stress-free environment for optimal production. Though the UK has some of the highest animal-welfare standards in the world, there are pressing concerns over maintaining this post-Brexit due to potentially cheaper low-welfare imports from outside Europe. Other constraints (e.g. high feed costs, environmental regulation, consumer concern over animal welfare and food security, long-term public-health concerns such as antimicrobial resistance) add to the pressure in this cost competitive sector, driving governments and consumers to demand greater supply-chain control. Unfortunately human monitoring involves many limitations, including contamination and farm-worker health risks, and provision of only limited frequency, resolution and fidelity of data. It is also a slow, costly and labour-intensive process. Automation offers the potential for continuous data capture, allowing more timely and effective intervention, improved animal welfare and reduced production costs.

**Robotics and autonomous systems for aquaculture**
Aquaculture production is already a vertically integrated and professional supply chain, but operates in an environment with a number of challenges that limit production, where sensors and robotic systems can play a role. Any systems deployed are naturally required to be more robust to extreme conditions and environments. The environment for aquaculture is often hostile and difficult to access by human operators with remote locations and inclement weather, with access only by small boat, leading to high operating costs and significant health and safety issues. The use of autonomous sensing and remote operation could significantly reduce the requirement for an on-site human presence making such facilities safer and easier to manage. Major challenges include environmental and health issues, such as algal blooms, sea lice and gill diseases. Could robots be used to monitor and offer a mode of treatment to control parasites, for example? Could robots assist in benign control of seals and other protected species? Are there autonomous means of maintenance for the required infrastructure, especially as the industry has started to move recently to “higher energy” sites? Could more precise understanding of the environment and behaviour of fish lead to better management control and therefore better productivity? These and many other challenges exist in the aquaculture sector, even those systems that are land-based.

## 1.5 DISADVANTAGED FARMS

The vast bulk of previous agri-robots projects have focussed on agribusiness style environments such as large flat monoculture fields, and controlled industrial scale indoor growing. In Britain, these environments are predominantly in the South and East, and their productivity has historically been a major reason for these areas' economic successes. In contrast, the North and West's geography contain more varied and hilly terrains, which has created a different type of farming based around smaller family farms, smaller vehicles and often more intensive manual work, especially dairy and sheep farms. In EU classification these are known as "disadvantaged" (and the more extreme cases as "severely disadvantaged") farms. Disadvantaged farms have largely been left behind by successive waves of automation including agri-robotics. However in the distant past they were once as productive as the agribusiness areas: the Peak District, for example, was farmed intensively during the Bronze age, and many hill farms and now moorlands have gradually fallen into disuse purely because large machines cannot navigate their terrains as human workers once did. Agri-robots designed specifically for these terrains could help make them economically viable.

## 1.6 NON-CONVENTIONAL CLOSED ('VERTICAL') FARMING

'Vertical farming' systems utilise indoor farming techniques and closed environments where all environmental factors, including nutrients, temperature, humidity and lighting, can be controlled [17]. The co-design of robotics with specialised sensing and crop genetics, where plants have been bred to take advantage of the closed environments, is now enabling these systems to move from niche low-volume markets, such as localised production of herbs for high-end restaurants, to mainstream reliable delivery of volume produce at 'Amazon' scales. Such large-area enclosed farming systems have been referred to previously as 'vertical farms' to emphasise the high density configuration of plants. In reality, robotic sensing and effector units, typically gantry-mounted, may be engineered to accommodate plants in any combination of vertical or horizontal modules to optimise the use of space. Thus, such systems are particularly well suited for high-density urban environments, where the production of crops with guaranteed growth times near the point-of-purchase enables minimisation of both waste, due to over-production, and the carbon footprint associated with long-distance transport chains. Rural economies may also gain benefits from access to locally derived and nutritionally advantageous fresh produce lines that have traditionally been cost-prohibitive or logistically impossible to deliver.

The required components for 'vertical farms' include temperature and humidity control, balanced crop-nutrient chemistry, process engineering of hydroponics or other growing media, semiconductor illumination, non-invasive sensing and robotics. Now these engineering elements are all realisable, the final ingredient required is the design of crops specifically for such systems, as opposed to their translation from field or polytunnel varieties. Within these controlled environments, crops can be bred specifically to maximise yield or other desired output traits, without the need to breed for other factors, e.g. resilience to specific pests, weeds, etc., as in conventional growing systems. Robotic systems would then allow novel output traits to be nurtured and bred into future produce, such as beneficial health and nutritional aspects, or to minimise the energy and waste in any downstream processing. A further benefit would be to deliver plants that are more compatible with autonomous plant care and harvesting.

While 'vertical' production systems may not be economically or environmentally competitive with the existing systems for the foreseeable future, the potential for such units within urban and industrial environments cannot be considered in isolation. As plants are potential sinks for effluents, heat and excess energy production then the rationale for such food production facilities needs to be considered alongside their potentially beneficial effects that would otherwise have to be dealt with via alternate, less sustainable means. An exemplar is the positioning of British Sugar Ltd. as the biggest UK tomato producer as a consequence of being a sink to the carbon dioxide release from the sugar processing and the exothermic reactions which promote tomato growth. By appropriate design of the automation and robotics within non-conventional vertical farming units, the economic arguments for installing such systems may be overlaid with their ability to solve issues within parallel sectors such as exploiting alternative waste and excess energy streams.

## 1.7 FOOD MANUFACTURING AND PROCESSING

While post-harvest activities are beyond the main focus of this white paper, we note that the need for new research and innovations across Agri-Food does not stop at the farm gate. For example, the meat sector presents particular challenges in productivity, due to the increasing difficulties in finding skilled workers with the qualifications needed to carry out meat cutting tasks. Robotisation of these jobs has thus become an important objective for companies looking to improve the safety and health of workers, as well as finding new solutions to mitigate the increasing production costs linked to current and future labour shortages. Collaborative robotics (cobots), where robots work together with humans, presents an alternative solution to help increase productivity, improve health and safety, and attract skilled workers and graduates to the food processing industry.

Agricultural robotics could also facilitate earlier labelling and tracking of food products throughout the manufacturing supply chain, bringing numerous benefits such as improved information for consumers on where their food is from and faster action to mitigate food safety issues. In turn, the food metadata could be fed back to the field operations to further improve the primary production.

In general, there may be many potential synergies between agricultural robotics and the downstream processing of agricultural products in the food chain, where whole supply chain efficiencies could be unlocked through future application of RAS technologies. Common challenges that span the whole Agri-Food supply chain include the demand for soft robotics; human-robot collaboration; safety, of both food products, people and other assets; automation of intra-logistics, from in-field transportation to packhouse to warehouse operations; sensing and image interpretation for analysis and manipulation of complex food products; and long-term autonomy, requiring the development of robust, fault-tolerant systems able to operate 24/7 in challenging field and factory environments.

## 1.8 ETHICAL ISSUES

Various ethical issues arise from the emergence of robotic technologies in agriculture, with growing concerns over the impact of AI technologies on employment across sectors. A generation ago, manual labourers in UK farms and food factories were predominantly British, while today's industry relies heavily on c. 65,000 migrant labourers. There is a similar pattern worldwide, with migrant labour replacing native workers across developed countries. In turn, these workers may wish their own children to achieve a higher standard of education and work in more skilled jobs, while the demographics of an ageing population further limits the supply of manual labour. The average age of a UK farmer is 58 years, while many agricultural jobs require high levels of intrinsic fitness, which is not necessarily compatible with the demographics of age. We see robotic automation not only as a means of performing the "dull, dirty and dangerous" jobs that people no longer wish or cannot do, but also as a creator of desirable and rewarding employment, enabling human jobs to move up the value chain and attracting skilled workers and graduates to Agri-Food.

There are ethical issues concerning the ownership of data, similar to other technological domains where a small number of companies own or control the majority of the information and potentially the infrastructure. In a similar fashion, a power asymmetry already exists between farmers and large agribusinesses [18], Smart Farming faces two extreme future scenarios: 1) closed proprietary systems where the farmer is part of a highly integrated supply chain, or 2) open systems in which all stakeholders are flexible in choosing technologies and business partners [19]. Therefore, developments such as open-source data and publicly-funded data collection networks should be considered. Attention will also need to be given to the security of data collection, ensuring that objective measurements are taken and that the data can be relied on for use in decision making [20]. Legal and ethical aspects of autonomous agricultural robots, including liability frameworks and re-use of robot-collected data are further discussed by Basu et al. [21].

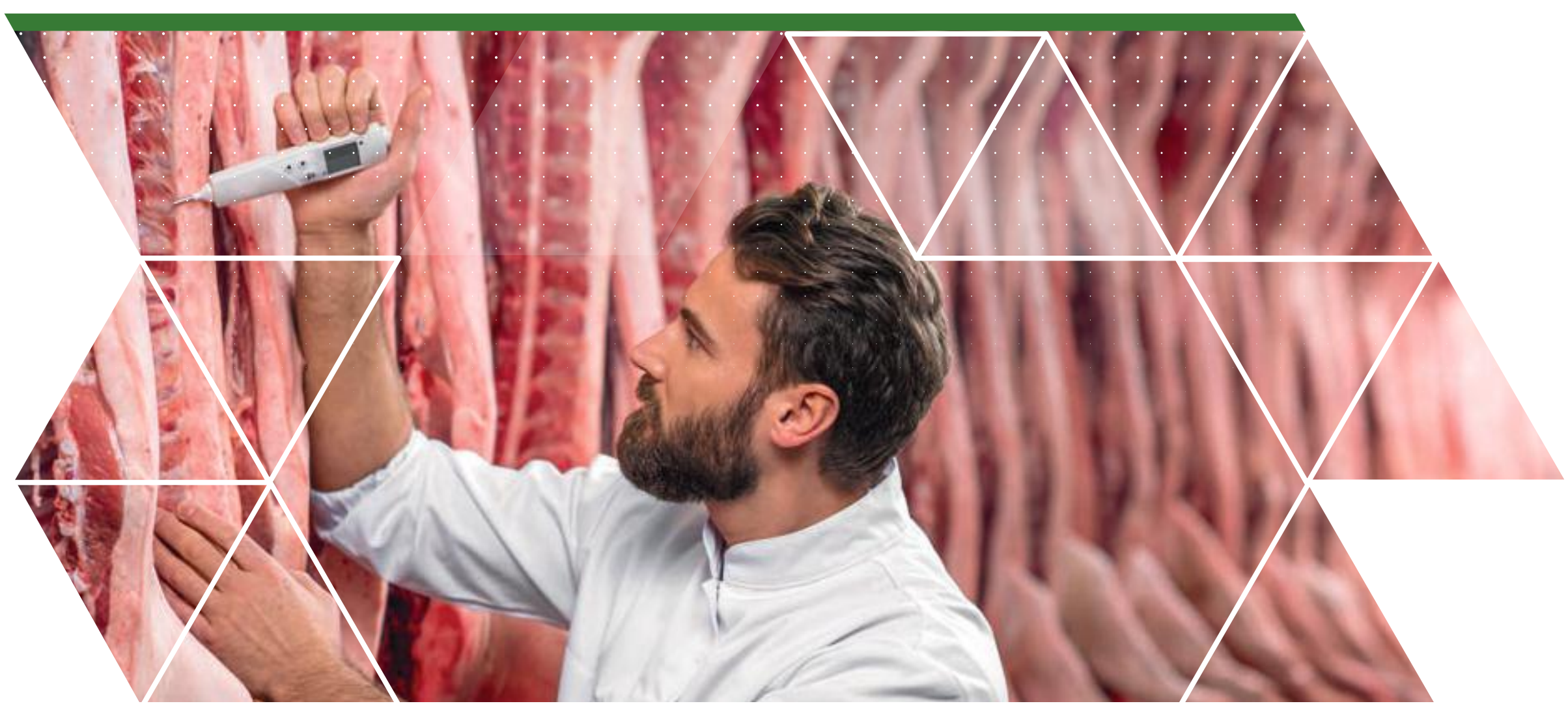

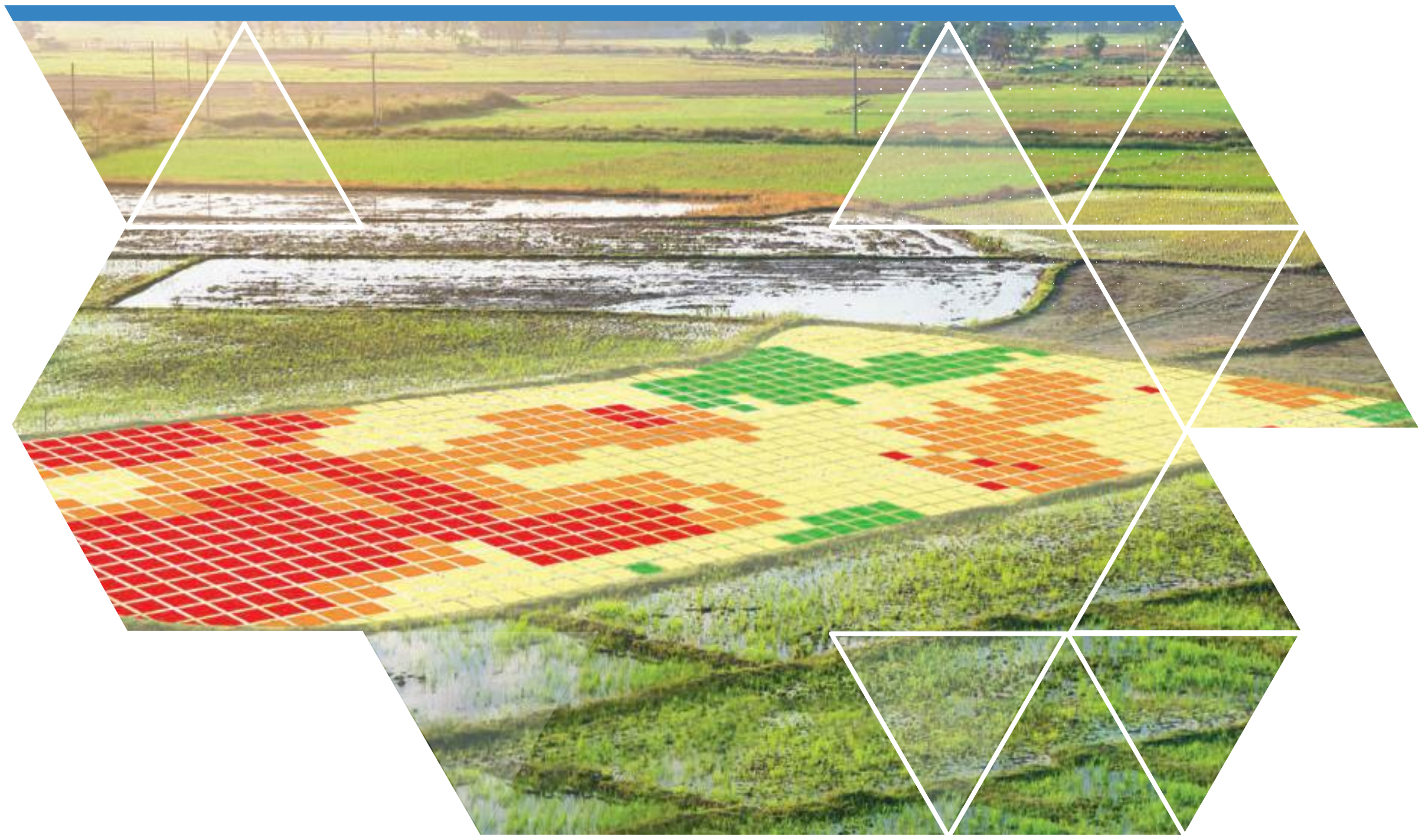

"The development of precision agriculture has been driven by the desire to better handle the spatial and temporal variability, e.g. in soil water-content or crop varieties, from farm-scale, down to field-scale, through to sub-field scale. The advent of autonomous system architectures gives us the opportunity to develop a new range of flexible agricultural equipment to reduce and target inputs in more effective ways."

# 2. TECHNOLOGICAL FOCUS

The recent focus of the agri-robotics community has been to identify applications where the automation of repetitive tasks is more efficient or effective than a traditional human or large machine approach [22, 23]. Research is needed into robotic platforms that can operate close to the crop (either on the ground or at elevation) and advanced manipulation, especially with interactive or tactile properties, e.g. for picking soft fruit. The use of heterogeneous "multi-modal" platforms that combine ground-based and aerial vehicles provides opportunities for targeted support and intelligence for the individual platforms, plus the ability for human operators to have an "eye in the sky" for observation and mission planning. Collaborative and cooperative behaviour becomes advantageous for large-scale arable and fruit crops as tasks can be performed in parallel, giving economies of scale. Land management is a specific issue of concern in the UK landscape, given the issues of fertilisation, water management and carbon content in the soil, so the use of advanced sensing and soil management using remote platforms including robotics will be increasingly important. Additionally, the use of robotics for livestock management is a specific opportunity for the deployment of autonomous platforms, as has already begun in automated milking stations, and with potential applications for raising animals in fields, barns, sheds and aquaculture, or fish farms.

## 2.1 RELATED AREAS

There is a plethora of related areas that are already using automation (such as in large parts of the industrial food production in the UK), and research is needed to investigate how this can be more tightly integrated into the agriculture industry. The food chain is managed using complex food production and software systems that rely on accurate data about all aspects of the location, quality and quantity of agricultural foodstuffs. Robotics and automation are already being used extensively in the processing side of the food industry, however this is not being leveraged to the same extent on the production side in the field. The application of large data sets in combination with remote sensing (as used in tracking raw materials for industrial production, e.g. with RFID tags) to optimise the quantity and quality of crops or livestock produced has the potential to revolutionise the UK agricultural sector. Technologies from related areas including the Internet of Things, Big Data and artificial intelligence can be used alongside autonomous systems technologies to automatically fuse and interpret collected data, assess crop status, and automatically plan effective and timely interventions in response to sudden events and the change of crop conditions (e.g. weather, diseases, pests).

## 2.2 TECHNOLOGY VISION

Our long-term technology vision encompasses a new generation of smart, flexible, robust, compliant, interconnected robotic systems working seamlessly alongside their human co-workers in farms and food factories. Teams of multi-modal, interoperable robotic systems will self-organise and coordinate their activities alongside and within existing Agri-Food systems. Electric farm and factory robots with interchangeable tools, including low-tillage solutions, novel soft robotic grasping technologies and sensors, will support the sustainable intensification of agriculture and drive manufacturing productivity throughout the food chain. Future agri-robotic systems will deploy artificial intelligence and machine learning techniques to increase their own productivity. Meanwhile, investigation of alternative systems for food production, including innovations from areas such as vertical farming, will further help to address the sustainable intensification of agriculture, while protecting the environment, food quality and health. A vital aspect of making this transition effective is the clear demonstration of economic benefits, which has always been the primary driver of change to the agricultural community.

**Facilitating the transition to automation**

While full automation is often hailed as the ultimate aim in technological development, and the future agriculture systems may look very different from those of today, only very few large companies can afford the disruption of full automation. So to achieve this long-term vision will require a gradual transition from the current farming practices, and most farmers will need technologies than can be introduced step by step, alongside and within their existing systems. Furthermore, while some emerging robotic technologies are already achieving or approaching the robustness and cost-effectiveness required for real-world deployment, other technologies are not yet at that stage. For example, soft fruit picking still requires fundamental research in sensing, manipulation and soft robotics. Thus, at least in the short-term, the collaboration of humans and robots is fundamental to increased productivity and food quality [24]. There are a number of comparatively low-cost platforms available now that are certified for use alongside human workers. Work is needed though to identify the nature of the robot-human interactions and joint workflows needed.

Thus, much of the research needed in the short- and medium-term should focus on facilitating the “transition to automation”, with mixed systems likely to dominate in the coming years, benefitting from collaboration between humans and robots, with combinations of electric, diesel and hybrid powered vehicles, as the required technologies for electrification mature and become ready for market. In the short-term, progress may depend on retrofitting of existing farm vehicles. For example, a human-driven tractor could tow a variety of robotic implements for different field operations such as selective harvesting or weeding. In the longer term, autonomous robotic vehicles will start to replace the legacy vehicles. This trajectory will also enable the UK vehicle and implement manufacturers to develop new products that span the transition from the current diesel-powered farm vehicles to the robotic farming systems of the future. The UK is well placed to implement these changes due to its strong automotive sector in industrial and agricultural vehicles, with extensive infrastructure already in place.

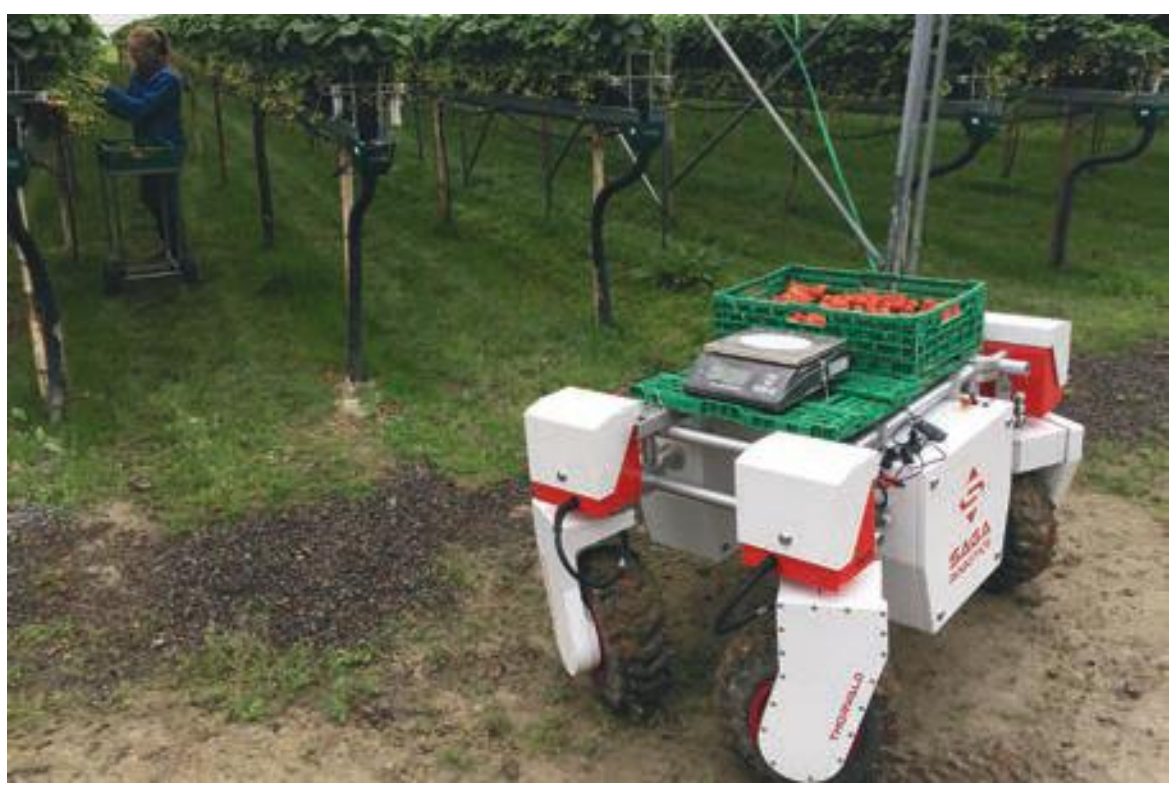

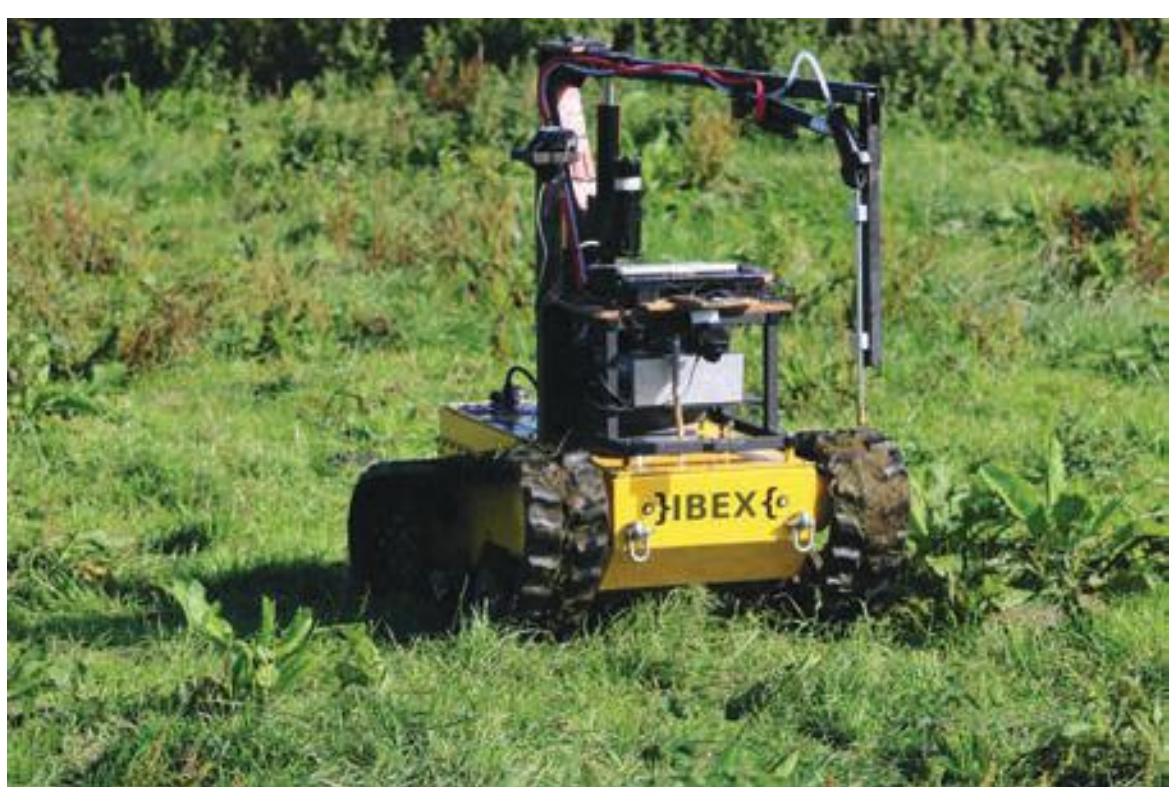

**Small, smart, interconnected, light machines**
One advantage of modern robotics is their ability to be built using low-cost, lightweight and smart components. Due to their prevalence in consumer electronics, such as mobile phones, gaming consoles and mobile computing (laptops, tablets etc), high quality cameras and embedded processors can be built in to many platforms at very low cost. New materials and fabrication techniques such as additive manufacturing and advanced composites are also making the manufacture and deployment of robotic platforms much cheaper and decoupled from a mainstream manufacturing process or supply chain. For example, a 3D printer located on a farm could be used to manufacture spare parts on demand, at very low cost, or even to improve the platform by adapting to the local conditions. Using collaborative and cooperative behaviour in a fleet of robots further provides the opportunity to spread tasks over multiple platforms and thereby reduce the damage caused by heavy conventional agricultural platforms on the soil or existing crops. The robotic fleet can also take advantage of multiple data sources to calibrate the task, reduce waste and focus on areas of greatest need, potentially reducing fertiliser costs and environmental impact.

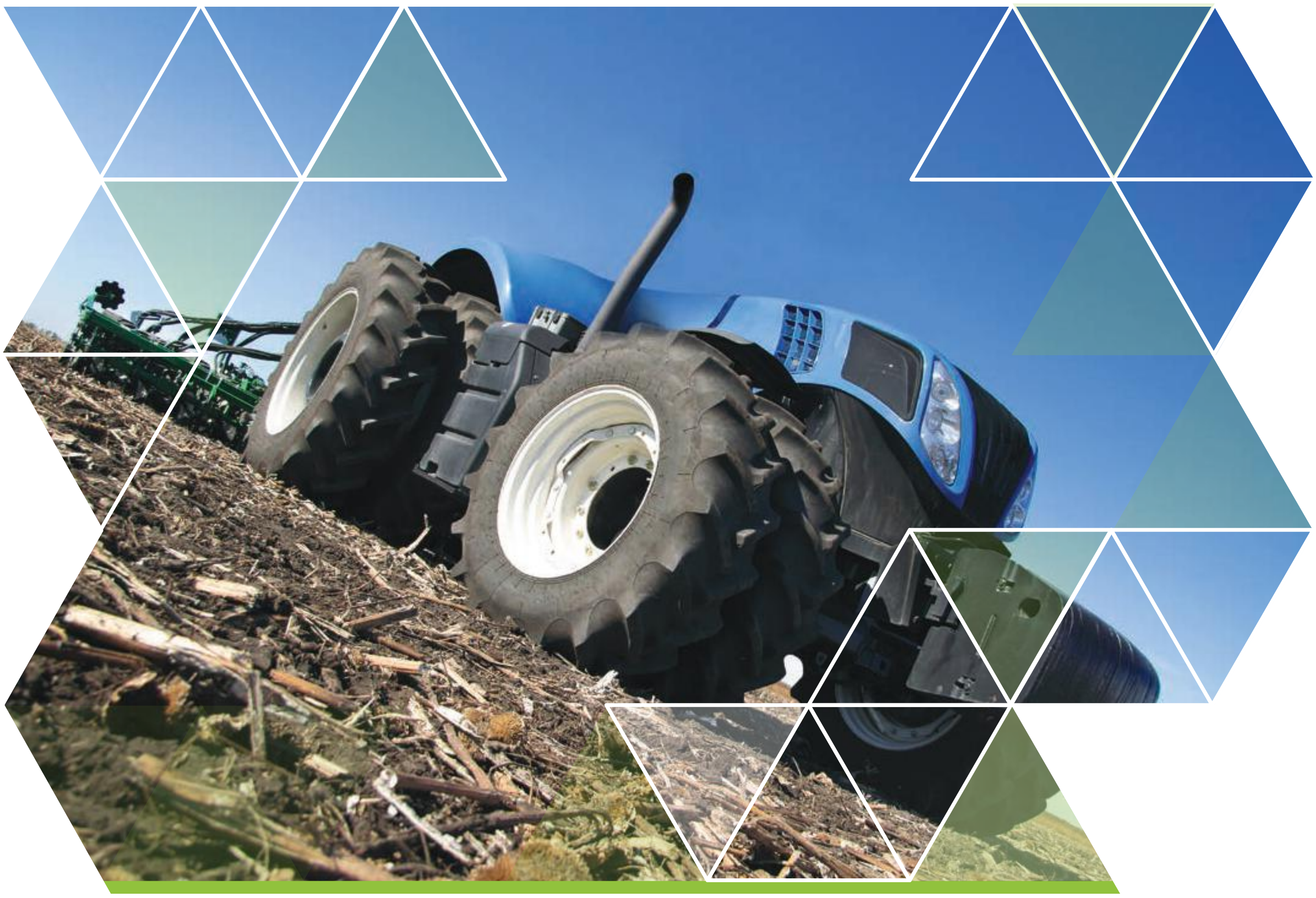

# 3. ENABLING TECHNOLOGIES FOR FUTURE ROBOTIC AGRICULTURE SYSTEMS

A wide range of technologies will enable the transition of agricultural robotics into the field. Some technologies will need to be developed specifically for agriculture, while other technologies already developed for other areas could be adapted to the agricultural domain, for example, autonomous vehicles, artificial intelligence and machine vision. Here we briefly review the current status, opportunities and benefits of various enabling technologies from hardware to software, multi-robot systems and human-robot systems.

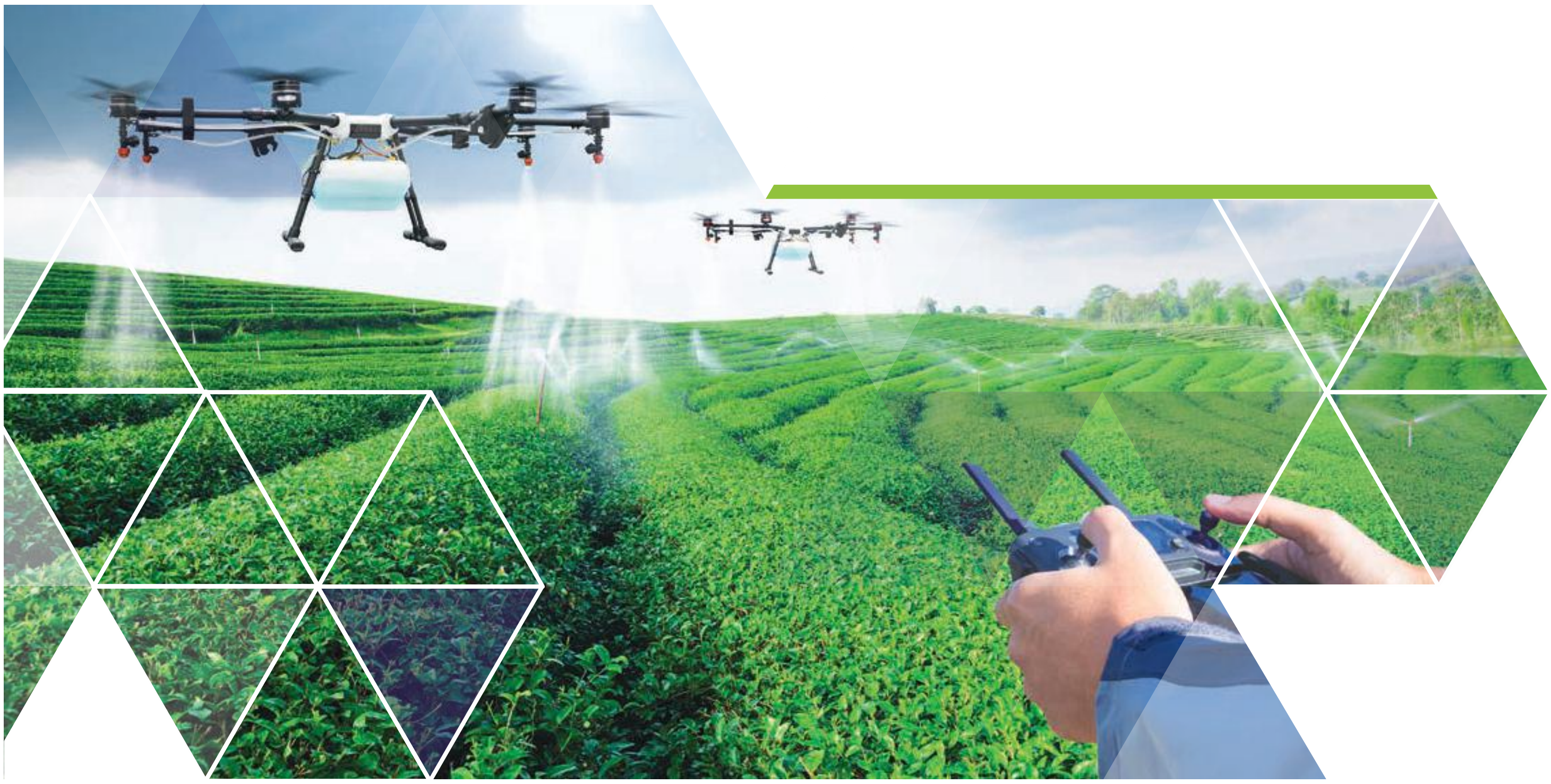

## 3.1 ROBOTIC PLATFORMS

Agricultural platforms can be divided into domain- and task-specific robots designed to perform a specific task on a given crop in a pre-defined domain, and generic platforms designed to perform several tasks in different domains. Both are likely to play important roles. Since farms in general have very different infrastructure, early robots may be able to operate only on a given farm and only to a limited extent across different farms. Similarly to current farm vehicles, we may see therefore a combination of robots adapted to a specific task and the emergence of multi-purpose robots able to carry out a multitude of different tasks, analogous to the myriad use cases of the modern tractor. A common challenge is that most current robotic platforms are not robust to real-world conditions such as mud, rain, fog, low and high temperatures. For example, most current manipulators are not equipped to deal with humidity in glasshouses.

### Mechatronics and electronics

The development of rapid prototyping techniques and low-cost processors have led to an explosion in the use of 3D printing and "maker" technology, raising the potential of low-cost robotic platforms for a variety of applications. The use of embedded software enables highly configurable and application-specific platforms that can use common hardware modules and be adapted to a variety of roles. While such approaches have been used extensively in UAVs and smaller-scale robots, there is much scope for the expansion of robotics in Agri-Food on a much wider scale. Issues that need to be addressed to migrate from prototypes to robust commercial platforms include robustness and reliability, power management (the platforms need to be able to operate all day, in some cases 24/7, for extended periods), usability (the platforms must be able to be used effectively by non-specialists), maintenance (e.g. self-diagnosis) and integration with mobile communications.

Further challenges include better characterisation of the mechanical properties of soil relevant to these robots, ruggedised platforms capable of operating in inclement weather, real-time sensing and control algorithms to adapt locomotion strategies to an ever-changing environment, and co-design of locomotion with other capabilities. For example, how does crop/fruit collection affect locomotion? What locomotion capability do we need to enable efficient sensing of crops?

### Locomotion

Agricultural robots need to move in challenging dynamic and semi-structured environments. Ground robots needs to travel on uneven, inhomogeneous, muddy soil, while aerial vehicles need to operate for long periods of time, in different weather conditions. Current agri-robots are mainly designed by borrowing technology from other sectors (e.g. drones) or as an add-on to existing platforms (e.g. autonomous tractors). As such, they may be not fully optimised for their tasks, or may retain some of the limitations of existing platforms.

UAVs can fly using multiple rotors or a fixed wing platform (with precision of location in the former and extended flight time in the latter), whereas ground platforms need to be able to locomote on rails and concrete floor in greenhouses, on gravel or grass in polytunnels, and in extremely muddy and difficult terrain in open fields [25]. We will therefore see a wide variety of robots being developed with different means to locomote. Compared to tractors, these robots are extremely lightweight, but as robots (or autonomous tractors) are to perform more energy-demanding task, the robots will also increase in size and weight. Most agricultural robots today run on batteries and electrical motors. Future developments will depend on how the battery technology evolves, but we will probably see both electric and combustion engines in the field for the foreseeable future.

A key aspect of any robotic platform is the impact of the weight and locomotion system on the ground and crops, and therefore different platforms have been used, including tracked and multiple wheeled robots. The platforms are also dependent on the required task, for example, heavy crop harvesting (such as volume arable or root vegetables) will need a heavier platform than soft fruit picking. Legged robots have the potential of minimising their footprint, while maximising the flexibility of locomotion (e.g. moving sideways or in narrow spaces between crops, etc). Their agility, combined with the possibility of carrying specialised sensors, may help unlock the full potential of precision agriculture.

### Manipulators

Manipulators will be needed for a range of tasks in future agriculture, replacing dexterous human labour, reducing costs and increasing quality, or performing operations more selectively than current larger machinery like slaughter harvesters. Work in this direction is ongoing, with soft grippers used for experimental work on selectively harvesting mushrooms, sweet peppers, tomatoes, raspberries and strawberries. Other applications such as broccoli harvesting can be performed with cutting tools, but will also require gentle handling and storage of the picked crop. In the open field, and for protected crops, there are complementary tasks to harvesting where manipulators can also play an important role. This includes mechanical weeding, precision spraying, and other forms of inspection and treatment. Manipulators will also be needed for the increased automation seen in food handling applications, such as large automated warehouses.

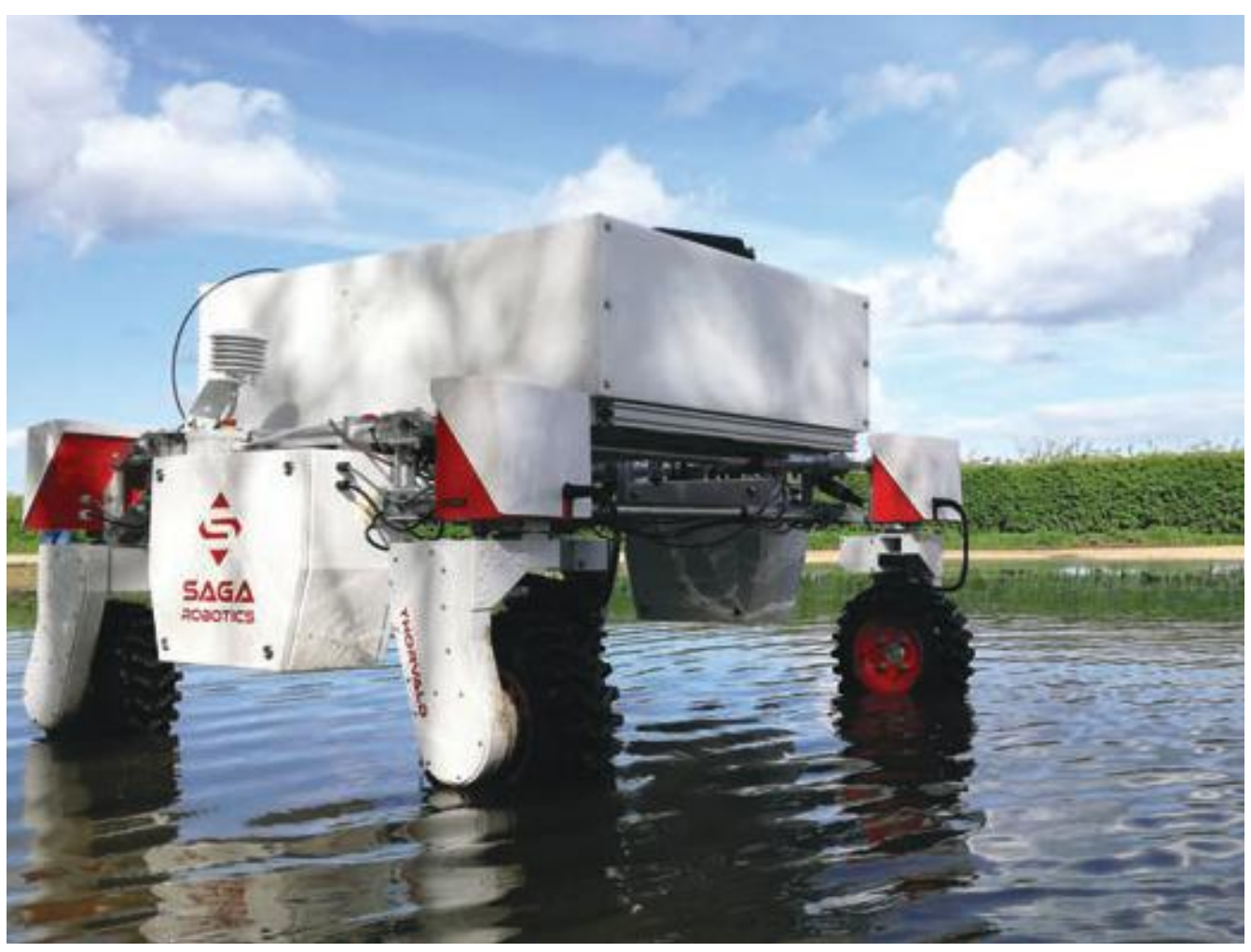

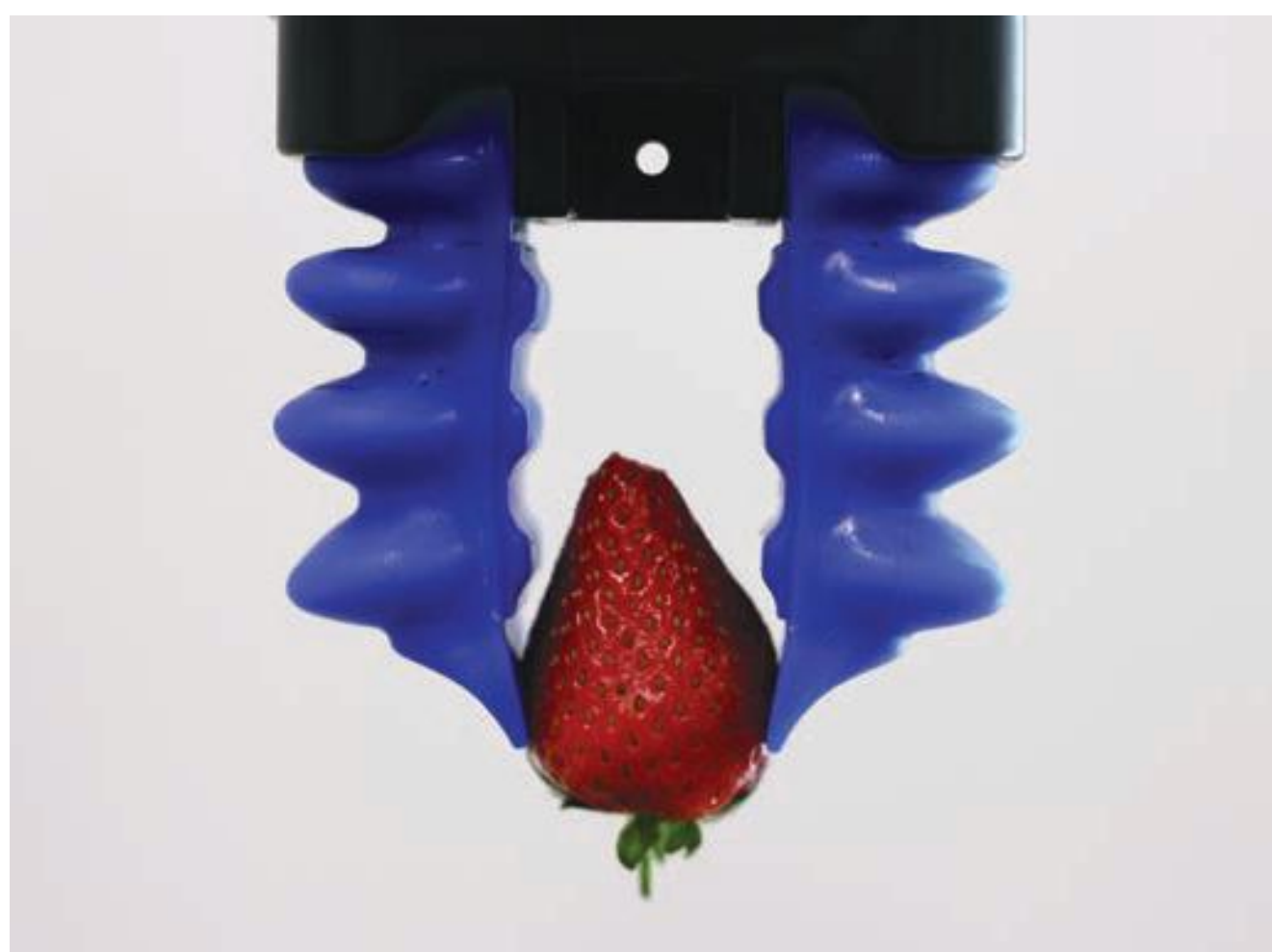

## 3.2 SENSING AND PERCEPTION

The integration of sensor systems within autonomous robotic systems offers the significant potential for new measurements that would otherwise be unobtainable. For example, current work addresses large area field mapping for bulk moisture by mobile robots, through the application of cosmic ray sensors adapted from the static COSMOS approaches [26]. Significant advances in satellite- or drone-based remote sensing capabilities open opportunities in monitoring crop growth status with unprecedented temporal and spatial resolutions while at an affordable cost. Many open source satellite datasets (e.g. from the European Space Agency [27]) are freely available for farmers. Robotic platforms further offer the possibility of forensic testing of soil with the geotagging and immediate results from sampling sensors (such as laser-induced breakdown spectroscopy), or secure collection of samples for later analysis in a systematic and uncontaminated manner. The use of compact robots and on-board secure collection systems will further enable a step change in the regulatory efficiency and reliability of land management systems using robots.

**Localisation and mapping**

The use of GPS navigation in agriculture has become almost ubiquitous with the deployment of RTK (Real Time Kinematics) allowing accuracy of centimetres for the automated positioning of large farm machinery such as tractors and combined harvesters. More recently, approaches using data manipulation of the GPS signal alone have shown promise to deliver equivalent accuracy without the cost of extra radio beacons. Accurate location data is not confined to unmanned vehicles with GPS, as precise localisation systems are available using visual fiducial markers and/or optical, acoustic or radio beacons, depending on the speed and accuracy required. Sensor information is also required in detecting objects and risks in field in order to ensure safe operation of robotic vehicles. To minimise damage to crops, the accuracy of relative positioning and navigation is more important than that of absolute navigation and position as provided by RTK GPS in many applications. For example, it is desirable to drive robotic vehicles to follow crop lines in accuracy of centimetres or follow the tracks left by previous tractor operations. Multi-modal systems based on a combination of GPS, INS, LiDAR, vision, etc have further potential for providing accurate and robust solutions, without requiring in-field infrastructure such as beacons.

Several attempts have been made to utilise seed and weed mapping concepts by passively recording their geospatial location using RTK GPS. Farming robots can be further equipped with pattern classification techniques that can predict the density and species of different weeds using computer vision. Other methods focus on a dense semantic weed classification in multispectral images captured by UAVs [28].

With the addition of advanced vision systems, including depth perception, scanning sensors such as LiDAR and artificial intelligence for decision making and classification, the concept of precision can be taken to another level. The ability offered by ground based robots to precisely control the location of scanning sensors, such as LiDAR, opens up the possibility to return quantified biomass estimates over whole crops as well as related phenotypic data, such as growth rates and morphology, through the integration of accurate location data with rangefinder scans using simultaneous localisation and mapping (SLAM) techniques. Similarly, robotic sensing platforms offer the potential for broad area analysis of insect pest or pollinator movement and their speciation, utilising 3D microphones alone or in combination with light backscatter measurements to enable daylight measurements of characteristic flight trajectories. Thematic maps can be built up for diseases, pests or weeds, which enable variable rate treatments, a key concept in precision agriculture.

**Crop monitoring**

The use of both land-based and aerial platforms can allow the third dimension to be accurately added to the management of crops using data fusion and SLAM techniques. This can be combined with virtual reality or augmented reality (VR/AR) systems to provide monitoring and intervention possibilities to an individual plant scale. Long-term data collection will further enable the modelling of crops over time, for example, tracking the development of the crop canopy, and thus improved prediction of future growth patterns.

Such ground and aerial robotic platforms offer additional prospects for enabling localised extremely high signal-to-noise, high resolution sensing that may not be achieved by passive remote (satellite) or semi-remote (rotary or fixed wing UAV) sensing technologies. At the simplest level, these robotic platforms offer the potential to extract close proximity (within 10s of millimetres) reflectance and transmission.

Multispectral Imaging (MSI) data helps compensate for the erroneous measures that occur due to the surface topology and orientation of individual crop tissues. At a more advanced level, the use of robotic manipulators to locate sensors around crops or livestock could enable responses to be tested and examined, through applying artificial stimuli. For example, through applying a focused beam of light at specific areas of crop tissue, and then modulating the spectrum and intensity, it is possible to drive photochemistries within specific parts of plants, e.g. stems, young leaves, senesced older leaves, etc., which can then be sensed via multispectral imaging. In this way significantly greater phenotype information may be recovered from across plants than could be achieved by passive fixed imaging detectors alone. Similarly, the cell structures and arrangements within fruits, vegetable and meats may be non-destructively examined in high resolution, e.g. for mapping subcutaneous bruising in fruits or fat ratios in meats. Nutrient and water stress of crops can be assessed by fusing MSI data with other data sources. Combining these assessments with crop growth models gives a better prediction of yield and loss, which leads to improved farming management and better food supply chain management.

### Robotic vision

Machine vision approaches offer significant opportunities for enabling autonomy of robotic systems in food production. Vision-based tasks for crop monitoring include phenotyping [29], classifying when individual plants are ready for harvest [30], and quality analysis [31], e.g. detecting the onset of diseases, all with high throughput data. Vision systems are also required for detection, segmentation, classification and tracking of objects such as fruits, plants, livestock, people, etc., and semantic segmentation of crops versus weeds [32, 33, 34], etc. to enable scene analysis (understanding "what" is "where" and "when") and safe operation of robotic systems in the field. Robotic vision in agriculture requires robustness to changes in illumination, weather conditions, image background and object appearance, e.g. as plants grow, while ensuring sufficient accuracy and real-time performance to support on-board decision making and vision-guided control of robotic systems. Active vision approaches, integrating next-best-view planning, may be needed to ensure that all the relevant information is available for robotic decision-making and control, e.g. where the fruit or harvestable part of a crop is occluded by leaves or weeds. Approaches based on analysis of 3D point clouds, e.g. derived from stereo imagery or RGB-D cameras, offer significant promise to achieve robust perception in challenging agricultural environments [30, 35].

Machine vision is already making an early impact in animal monitoring, e.g. for weight estimation, body condition monitoring [36] and illness detection [37] in pigs, cattle and poultry. Individual animal identification, e.g. using facial recognition techniques adapted from work in human facial biometrics [38], will allow more targeted precision care and timely interventions for individual animals, thereby ensuring their healthcare and wellbeing as well as optimising farm production.

Robotic vision often depends closely on machine learning from real-world datasets, with approaches such as deep neural networks [39, 29, 40] gaining traction and further raising the possibility for robots to share their knowledge by learning from Big Data. An open challenge in robotic vision and machine perception for robotic agriculture is to enable open-ended learning, facilitating adaptation to seasonal changes, new emerging diseases and pests, new crop varieties, etc. Most existing work considers only the initial training phase prior to deployment of a robot vision system, but not the ongoing adaptation of the learned models during long-term operation. The development of user interfaces for "ground truthing" and semi-supervised learning in robotic vision systems for agriculture is also an open challenge.

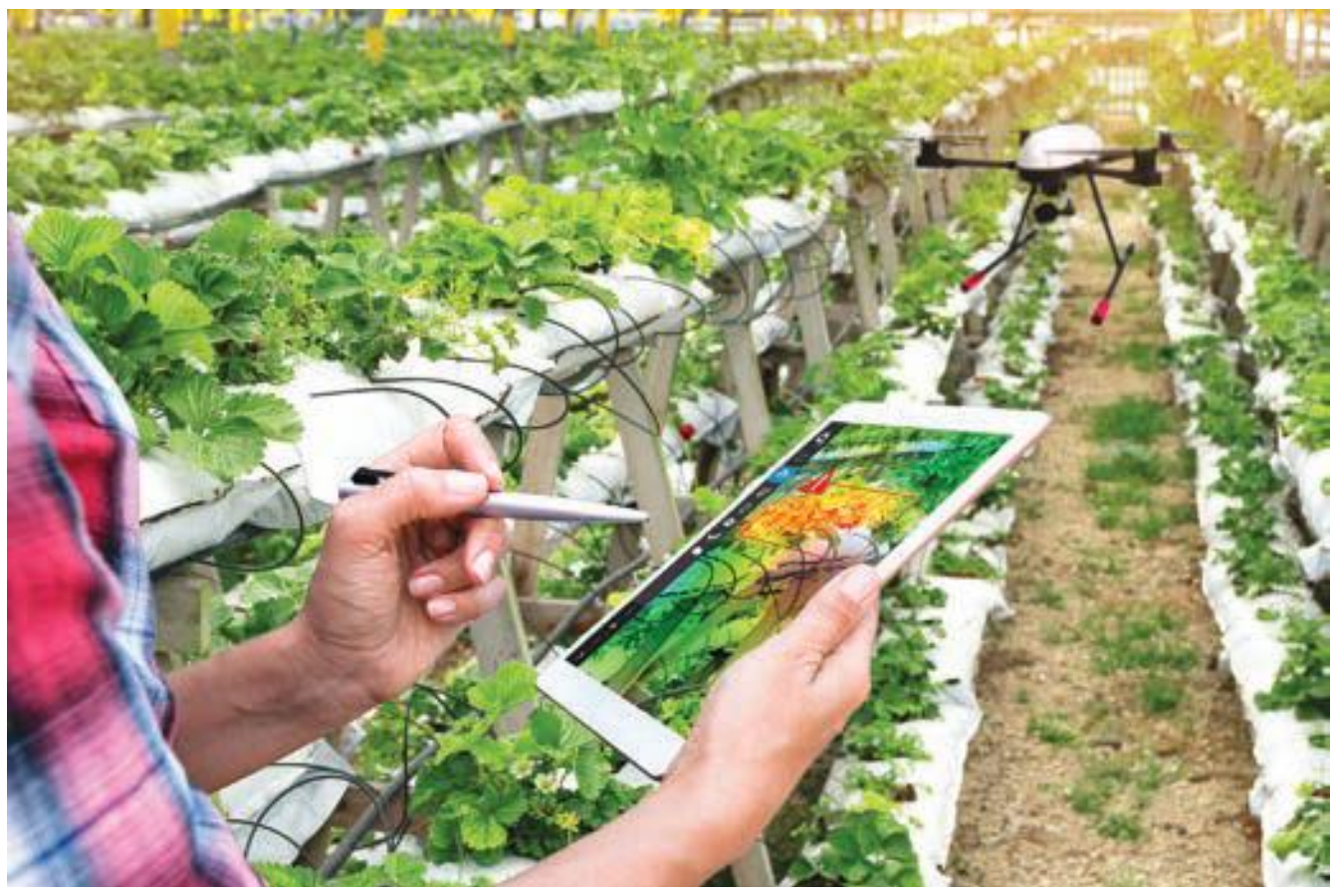

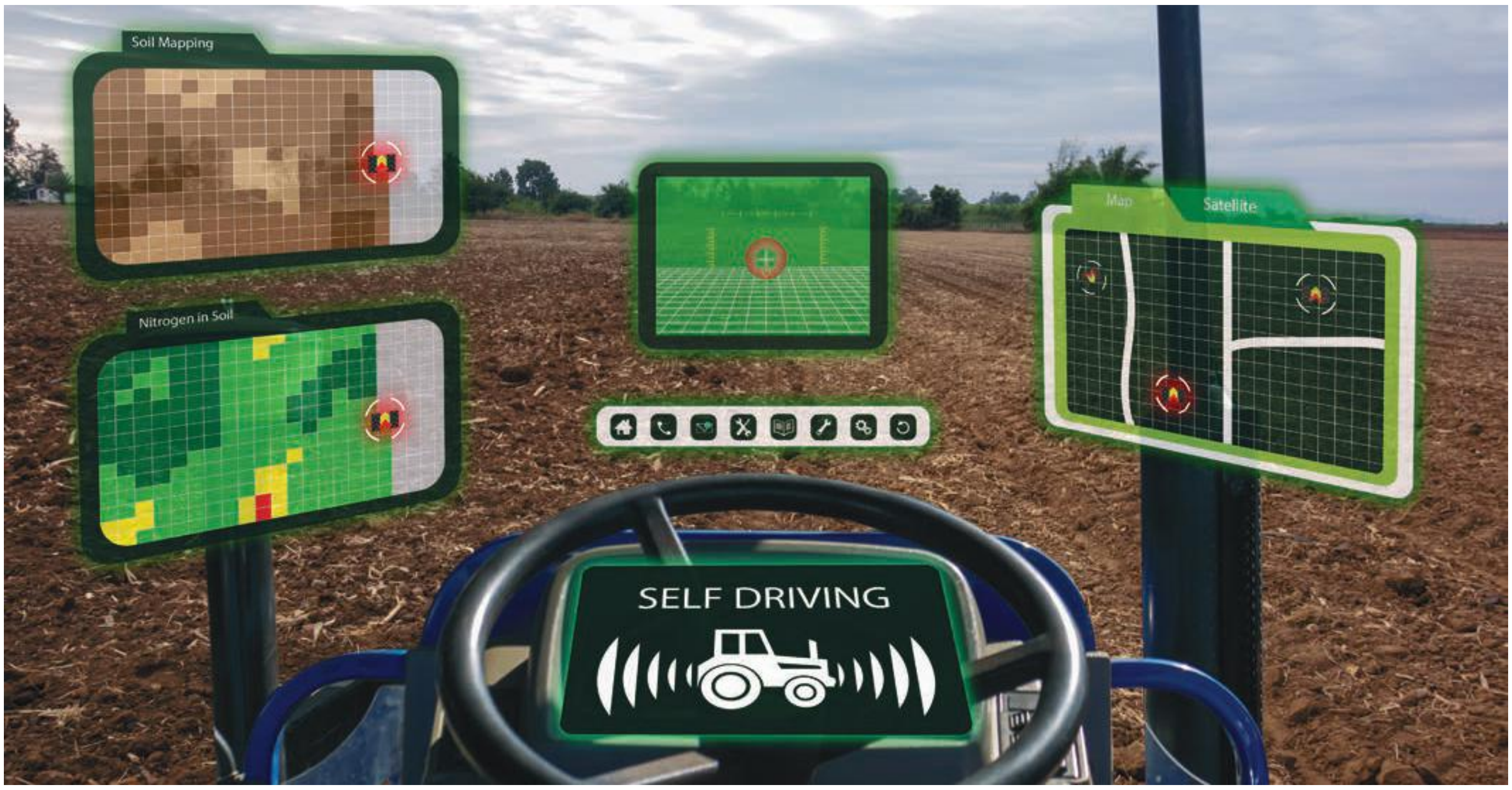

## 3.3 PLANNING AND COORDINATION

The true potential of robotics in agriculture will be harnessed when different types of robots and autonomous systems are brought together in a systemic approach. For example, UAVs are an excellent platform for environment monitoring, but with limited payloads and operational durability they are constrained when it comes to delivery of intervention or treatments on a larger scale. Hence, ground and airborne vehicles need to be integrated into heterogeneous fleets, coordinated either centrally or in a distributed fashion.

Planning, scheduling and coordination are fundamental to the control of multi-robot systems on the farm, and more generally for increasing the level of automation in agriculture and farming. For example, intelligent irrigation systems can respond to the change of weather conditions and crop growth status to automatically optimise the irrigation strategy so as to reduce the use of fresh water without loss of yields. The optimised strategy (e.g. when, where and the amount of water) is then implemented by computer-controlled irrigation equipment.

Such coordinated fleets also pose requirements for in-field communication infrastructure, such as Wi-Fi meshes, WiMAX ad-hoc networks, 5G approaches or other proprietary peer-to-peer communication methods deployed in field. On a larger scale, the heterogeneous fleets deployed in-field can also include collaborating humans sharing the working environment with their robotic counterparts, giving rise to interaction and communication requirements between the robots and the human operator and workers in this context. Example applications include in-field logistics, where vehicles need to be scheduled for area coverage and routing problems.

More generally, holistic approaches to fleet management are required, which fully integrate component methods for goal allocation, motion planning, coordination and control [41]. These sub-problems have so far largely been studied in isolation, so basic research on integration and scaling to real-world scenarios is required. Aspects of swarm robotics could potentially be applied to fleet management systems in agriculture, as in the EU-funded ECHORD++ projects SAGA and MARS [42]. To enable robot-human collaboration, the fleets also need to be aware of the presence of humans and to predict likely human actions in order to anticipate potential collisions and ensure safety. In return, the motion of robotic systems needs to be legible to humans, to facilitate acceptance by and cooperation with their human counterparts.

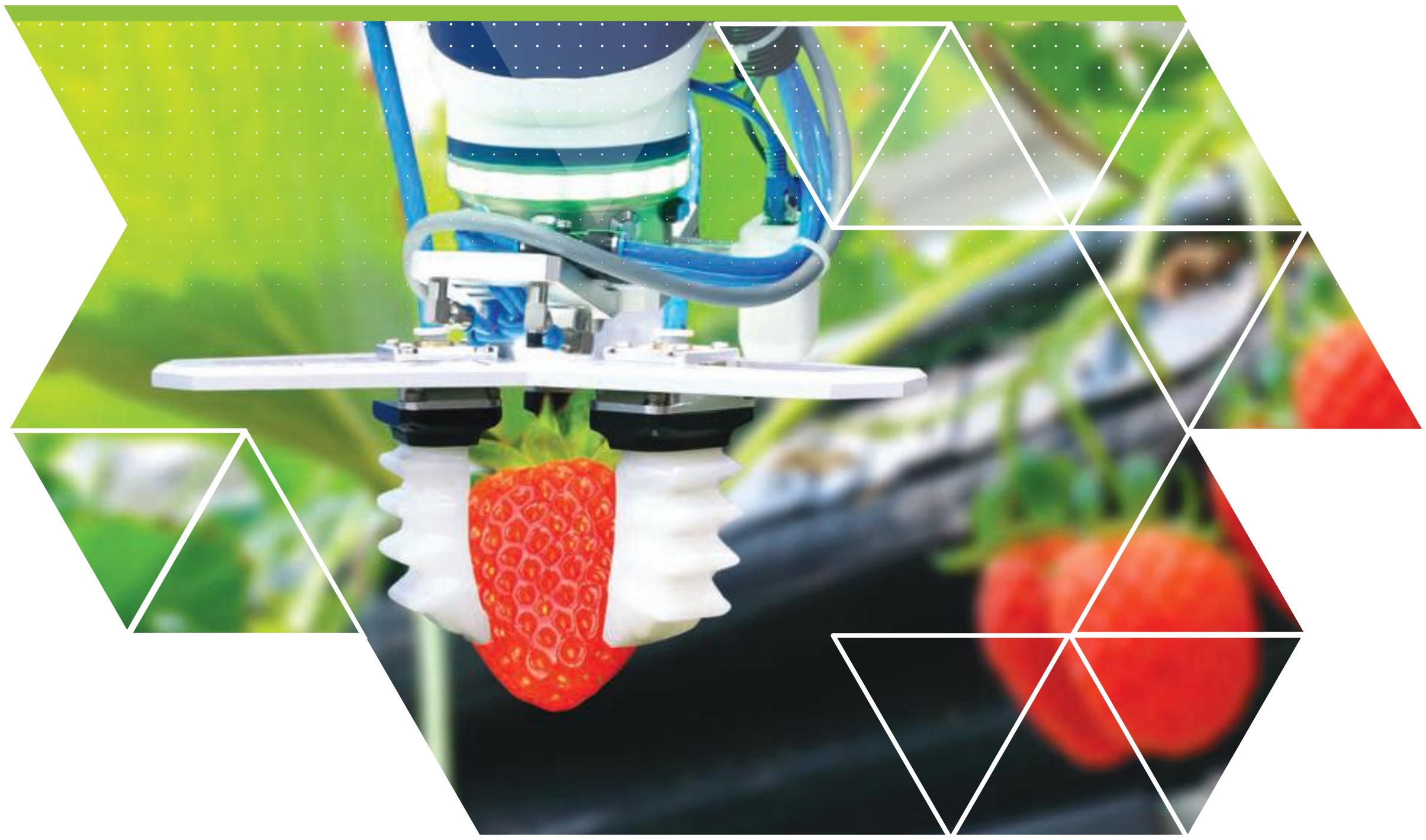

## 3.4 MANIPULATION

Automated manipulation and grasping of food items presents a series of unique challenges compared to other sectors. These include significant natural size and shape variations between examples of the same product, heterogeneous positioning of products (e.g. during harvesting) and the fragile nature of food products. Some areas of food harvesting have been successfully automated but these solutions are best suited to situations where the entire content of a field becomes ready for harvest at the same time, e.g. grains or root vegetables. If plants fruit over an extended period of time with only some ready to harvest at any particular time (e.g. tomatoes) automation struggles. This is because discrete items must be harvested individually without disturbing those around them and, due to the dexterity, advanced perception and decision-making required, human labour is still widely used.

Soft robotics [43, 44] is expected to play an important role. Soft end-effectors and grippers are needed for gently handling soft fruit and vegetables, such as soft robot hands for lettuce harvesting and suction devices for picking apples. Robots are increasingly made softer also on the actuator/joint level. Whereas stiff robot arms are suitable for blind operation in a factory environment, an agricultural manipulator requires sensorimotor coordination to achieve its task. Some tasks also require the right amount of force to be applied, dictating a force-based rather than position-based approach to control. In general, grasping and manipulation applications in Agri-Food require robustness to the unforeseen, while maintaining their ability to actuate with precision. One way to achieve this is through variable-stiffness actuators [45], which incorporate elastic structures, much like humans.

The development of compliant manipulators and grippers will in turn transform and simplify the design of agricultural robots by reducing the need for complex visual and tactile sensors. For this potential to be fully unlocked, novel design and control techniques need to be developed. Grasp planning is also a significant challenge. The most common approach is to use vision systems to locate products and use this to direct the grasp. However, this approach can fail if the object to be grasped is partially obscured by other products or foliage. Vision alone provides only limited data about an object during grasping and picking; human operators also use tactile feedback to adjust their action as a product is grasped to ensure it is picked successfully.

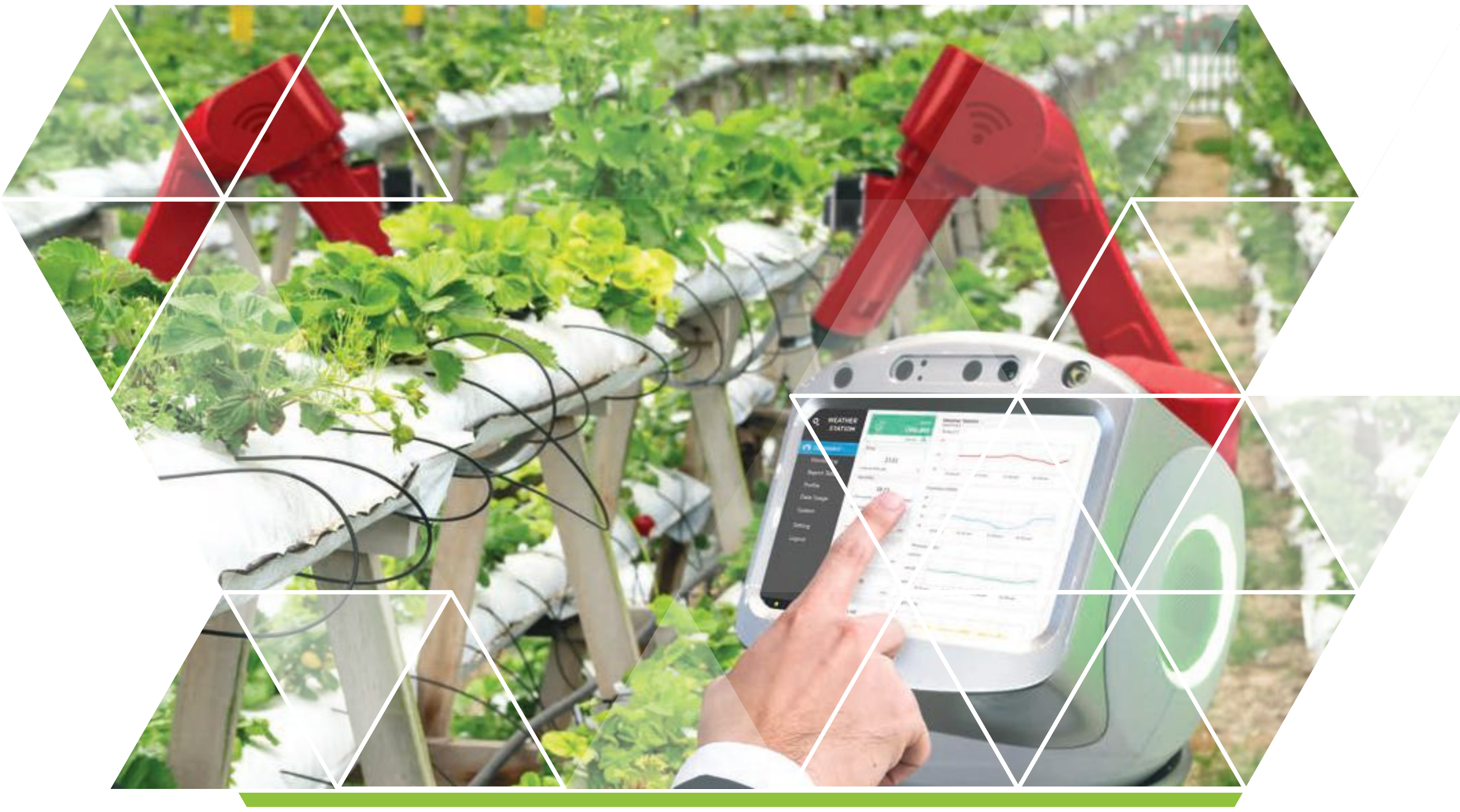

## 3.5 HUMAN-ROBOT INTERACTION

The challenges for interaction range from domain-independent aspects such as intuitive designs, immersive displays (e.g. Virtual and Augmented Reality) and tactile feedback, to very specific challenges stemming from the in-field conditions. Examples include the design of suitable interaction devices that are operable under harsh conditions, with constrained dexterity and precision of the operators, e.g. workers wearing gloves or having wet and muddy hands, or to guarantee the safety of often large and heavy semi-autonomous machinery in an environment shared with human workers. In contrast to robots in factories, where working areas can be fenced off when a robot is in operation, agricultural robots are limited by the absence of safety infrastructure in the fields, and require new innovative solutions.

### Human-robot collaboration

Robots closely collaborating with humans (so-called cobots [46]) are delivering real step changes in many industrial sectors, and are anticipated to be vital to automation in agriculture. Use cases range from farm in-field logistics (transportation), where efficient and safe hand-over of goods and produce needs to be facilitated, to applications enhancing animal and crop welfare by means of integrated monitoring and intervention delivery. An illustrative example is the RASberry project at the University of Lincoln, where human pickers of strawberries are supported by mobile robots acting as transporters.

While some tasks for cobots require physical interaction between robots and humans, in other areas robots can act as a mediator or provide a remote presence for agronomists and farmers. Therefore a focus on intuitive and ergonomically appropriate interfaces and interaction design is needed. Concepts of shared autonomy and control, allowing operators to exercise control from remote locations over a potentially heterogeneous (ground, air, water) fleet of semi-autonomous robots, will also be important. As the technology matures, and in particular for safety-critical tasks, various levels of shared autonomy will be seen, where the human operator guides the high-level execution, while the robotic system performs the required sensorimotor coordination on the ground. The fan-out [47], or number of robots a human can control simultaneously, will help drive the mixture of human supervisors and robot agents in such a paradigm.

### Safe human-robot interaction

By relying on humans as supervisors, the autonomy levels, and associated risks and design complexities, can also be improved. Human supervision will be a vital safety factor for most agri-robotic systems for the foreseeable future, while the technology develops towards higher levels of autonomy. The robotic systems will also be learning and adapting to task and farm-specific constraints. Human and robot collaborators will therefore likely be mutually adapting to each other, in order to maximise performance.

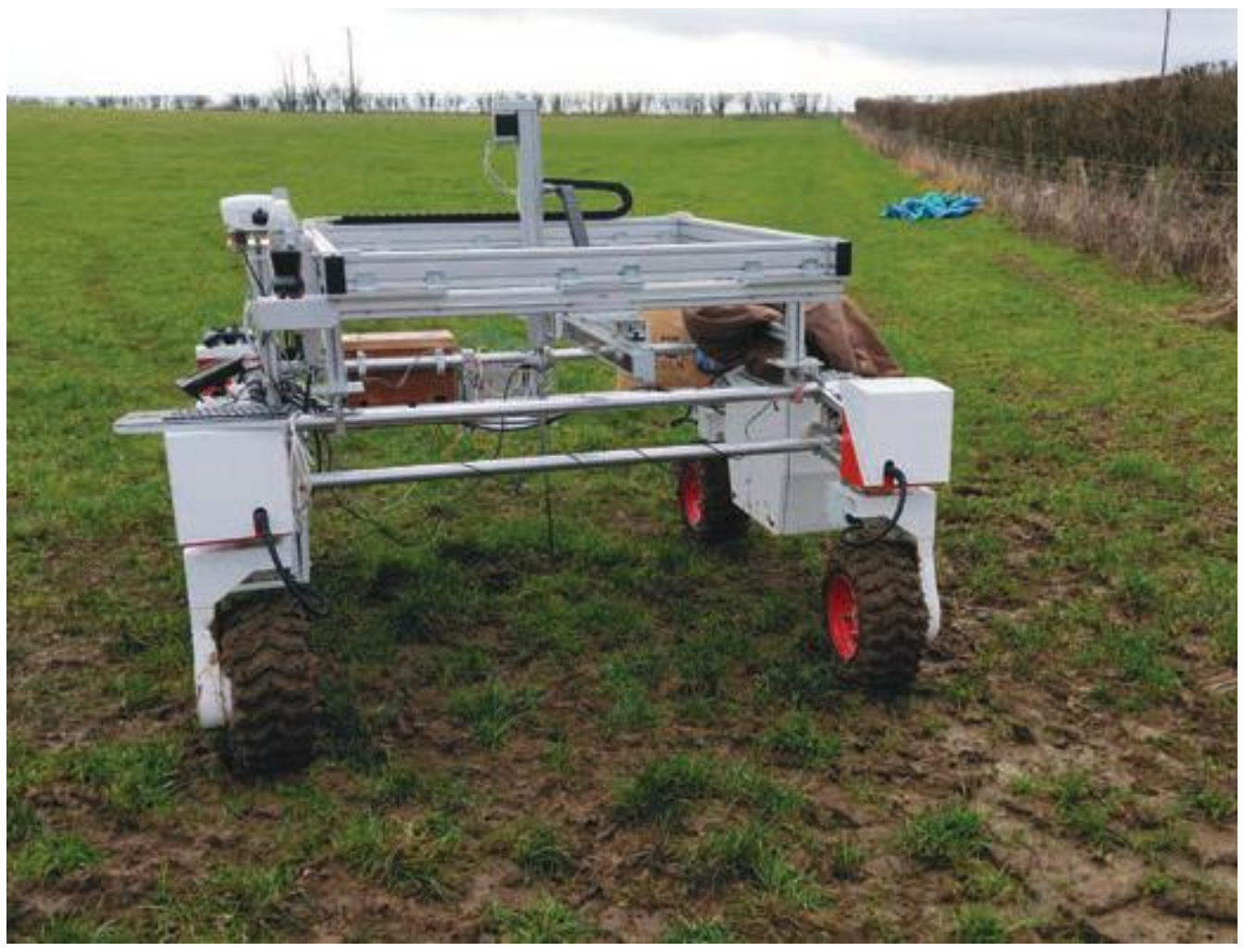

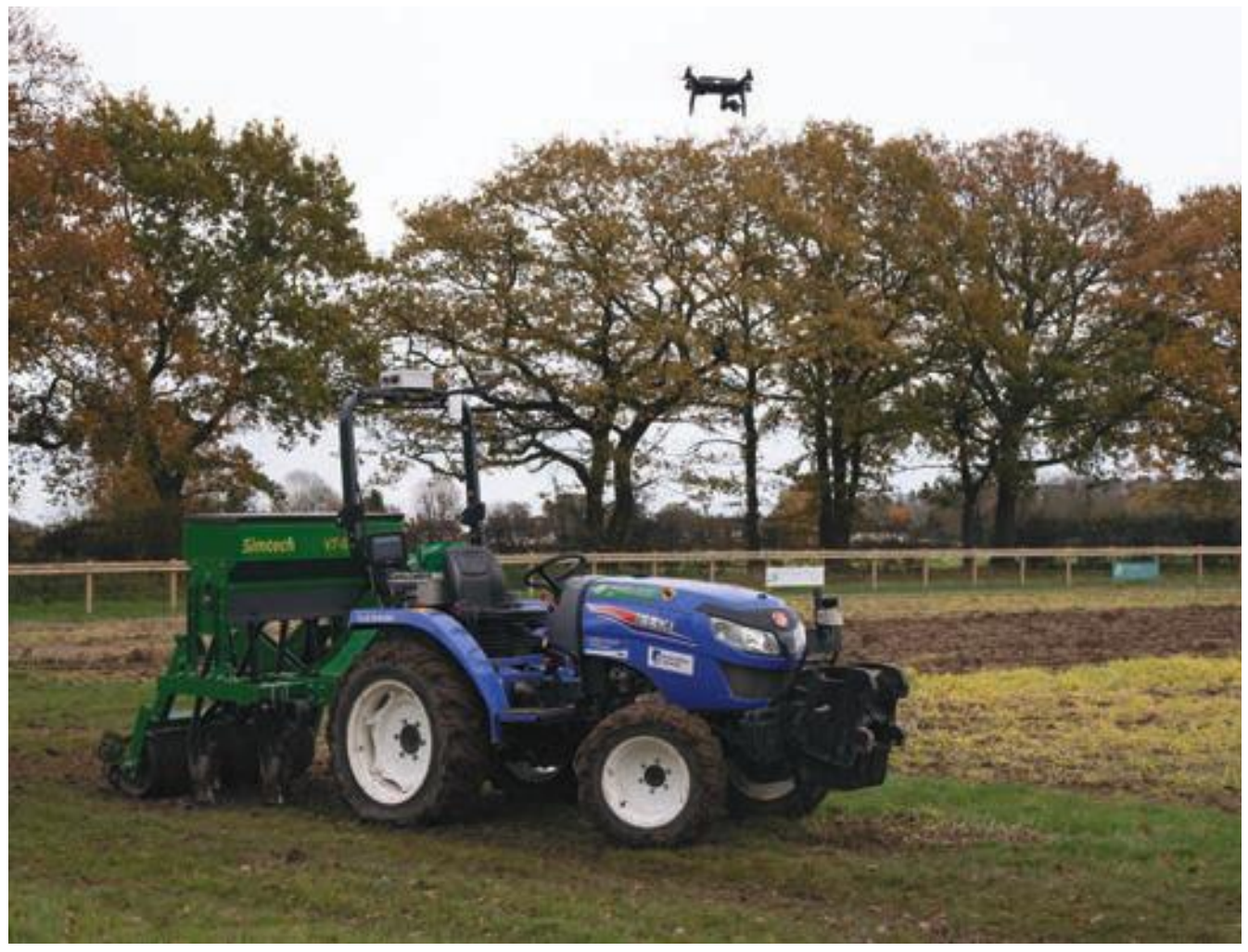

Approaches to safe physical Human Robot Interaction (pHRI) [48, 49] include supervisory systems to monitor the interaction and adjust the behaviour of the robot if an unsafe situation is identified. This typically involves slowing, or completely stopping the robot, to prevent accidents. However, this approach can significantly reduce productivity as the robot is not working to its full potential. Current research aims to improve on this approach by allowing robots to identify and predict unsafe situations, and then to adapt and adjust their operation to continue the task in a manner that allows both productivity and safety to be maintained [50]. A further approach to ensuring safe pHRI is to design robot systems which are inherently safe, meaning that if collisions occur between human and machine, injury will not result. The aim is to replicate the safe interaction that occurs when multiple people work collaboratively. This requires a change away from heavy, rigid and high inertia robots to systems which are more akin to biological creatures. Again this is a challenge that the new field of soft robotics may be able to address.

## 3.6 LEARNING AND ADAPTATION

Artificial intelligence technologies, especially in machine learning, are expected to play a major role in most of the above technology areas, and will be essential enablers for agricultural robots. Agricultural environments are subject to changes throughout the lifetime of a robotic system. For example, there may be new crop varieties, weeds, pests, diseases, treatments, legislation, climate change, etc., as well as new implements and robotic technologies. In AI terms this means dealing with an open world, so techniques to enable adaptation during operation rather than at the design phase will be crucial. Techniques that allow robots to learn from experience include reinforcement learning, learning from demonstration, and transfer learning to exploit prior knowledge, e.g. from another domain or task. Ongoing research is investigating deep learning methods [40], especially in perception-related tasks involving the interpretation of sensor data, including recognition and segmentation tasks in automated weeding and fruit picking. Robots will also need to leverage human knowledge, especially when facing situations that were not foreseen at design time. This additional input might be given by end-users, maintainers, and/or domain experts. It might also be provided through direct control (i.e. teleoperation), natural interaction (e.g. via language or gestures) or by the means of labelled examples and data sets. These developments will link naturally into the use of Big Data in smart farming [19], alongside the use of satellite imaging, UAVs and ground robots for more localised and richer, multimodal data collection. These developments coupled with cloud-based storage will create an abundance of information that could potentially be utilised for smart planning and control of agriculture. An important requirement is the standardisation of data to ease the exchange between robots, domains, farms, countries and companies.

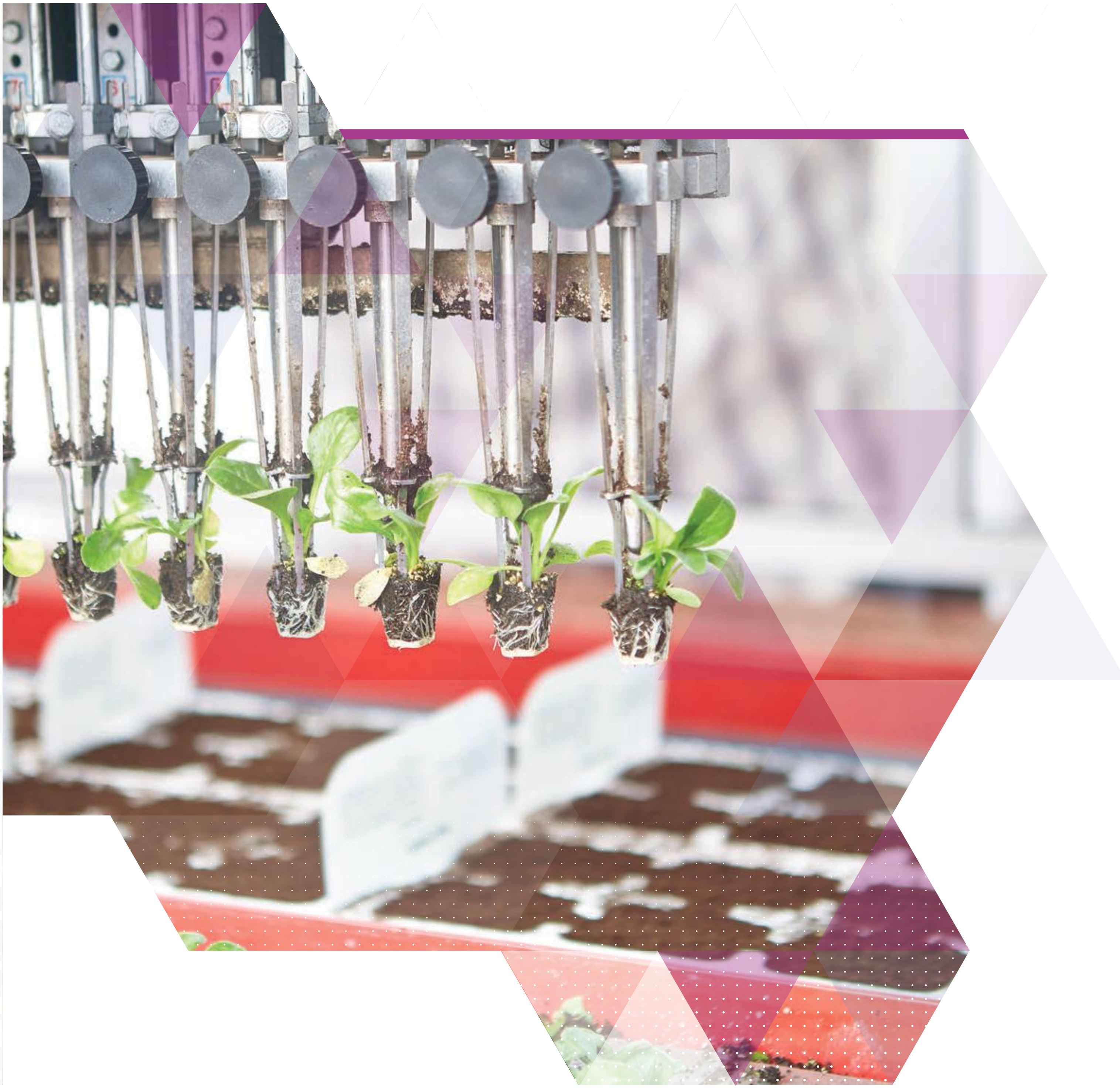

# 4. THE CHALLENGES

RAS technologies have the potential to revolutionise all agricultural sectors. The nature of their contribution will differ across agricultural types, i.e. crops, livestock and aquaculture, and from phenotyping through to primary production activities. Across this spectrum there are many ways that robots could contribute, both economically (e.g. growing and harvesting more efficiently and cheaply), ecologically (e.g. reducing and eliminating the use of chemicals, while helping to maintain soil health) and ethically (e.g. increasing animal welfare via monitoring and timely intervention). In parallel, robotics may enable automation in the care of livestock and aquaculture or alternative growing systems, such as ‘vertical’ (protected-environment) farming or agroforestry systems combining agriculture and forestry on the same land.

The technology challenge areas for agri-robotics tend to divide into two classes: (1) Breeding / Phenotyping and (2) Farming / Primary Production. There is then a whole additional sector of post-harvest Agri-Food activities for robotics, which fall outside the remit of this white paper. Taking conventional terrestrial arable agriculture as a farming exemplar, the challenges can be illustrated as follows. Direct parallels may then be extrapolated for livestock and aquaculture as well as non-conventional farming systems, e.g. organic or ‘vertical’, however for the purposes of brevity these have been excluded from the narrative below.

**Phenotyping**

- Laboratory: The identification and selection of genetics with beneficial abiotic (e.g. drought or saline tolerance) or biotic (e.g. fungal viral or bacterial disease resistance) input traits and complementary output traits (e.g. nutrient-to-biomass conversion, shelf-life, flavour or nutritional qualities) is conventionally undertaken within controlled laboratory environments. These breeding activities have seen a degree of robotic integration in recent years to reduce the reliance on manual intervention, but the cost and complexity of implementation, as well as the questionable reliability and technology-readiness of the current systems, has limited their uptake.
- Field: While laboratory phenotypic screens may be important for identifying beneficial breeding lines to cross, they are only a proxy for the primary goal of determining how that crop may thrive in real-world conditions. Robotics opens up the potential for mass direct in-field phenotyping of crops under true farm conditions [51, 52]. Such uncontrolled ‘non-laboratory’ systems raise significant challenges over the singular identification of the specific trait that resulted in a beneficial phenotypic response. The robotic capability for repetitive and detailed assessment of the environment of individual plants opens up the potential for a paradigm shift in the development of agri-genetics.

**Crop Management**

- Establishment and Seeding: Ploughing is one of the most important primary cultivation processes, and involves the inversion or mixing of topsoil to prepare a suitable seed bed. Currently modern agriculture uses a huge amount of energy in ploughing: it is estimated that 80%-90% of the energy in traditional cultivation is used to repair the damage done by large tractors. Small, smart, electric robots provide an alternative solution, by avoiding excessive compaction of the soil in the first place, and performing micro-tillage using on-board implements. Nutrients could also be better targeted to the local environment of individual seeds using precision approaches. Seed placement and mapping could be further automated to optimise the density and seeding pattern with respect to the requirements for air, light, nutrients and ground moisture of the individual crop plants. Robotics will also have an important role to play in managing the inputs to primary production, including both monitoring and interventions, particularly for soil [53] and water [54].
- Crop Care: One of the main operations in crop management is scouting to collect timely and accurate information. Autonomous robots carrying a range of sensors to assess crop health and status could thus assist in cost-effective data collection. Both aerial and ground-based platforms, or their combination [55], could be utilised. Fusing data collected by different devices or obtained from sources with a wide range of temporal and spatial resolutions and automatically interpreting data impose a number of interesting research challenges. Weed mapping involves recording the position and density (biomass) of different weeds species using machine vision. The resulting weed map can be further interpreted into a treatment map. Robotic weeding is an active area of current research, investigating alternative methods to kill, remove or retard unwanted plants without causing damaging to the crop. Intra-row weeding is more difficult than inter-row weeding, as it requires precise positioning of the crop plant. Alternative methods for

THE OPPORTUNITIES FOR ROBOTICS AND AUTONOMOUS SYSTEMS IN AGRICULTURE

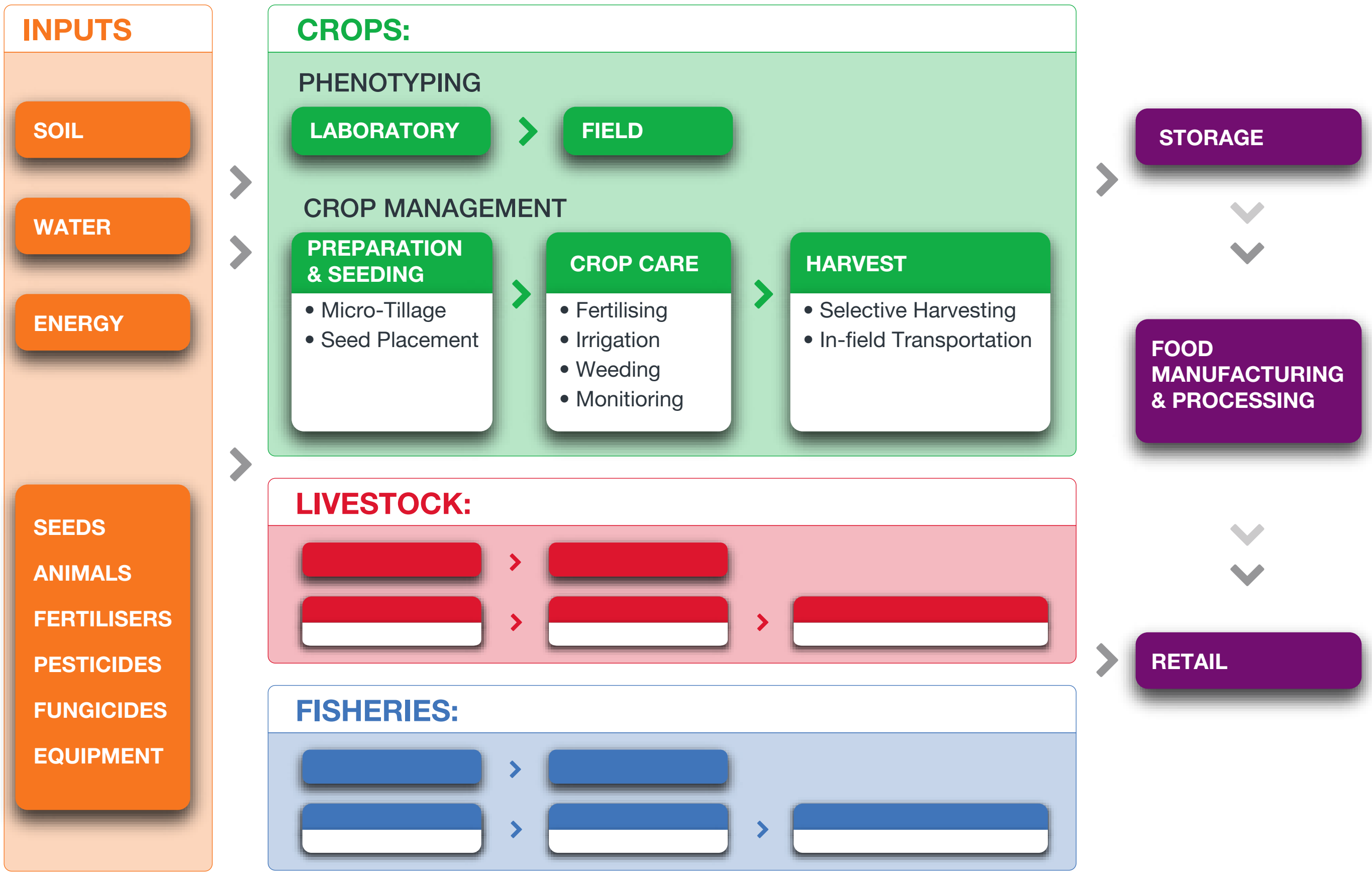

weed control include vision-guided mechanical weeding, selective (micro-) spraying and laser weeding. Irrigation is another area where robots could assist in targeting water in the right place at the right time. Pre-harvest assessment and yield forecasting [56] by robotic sensory systems will further assist in choosing the right time to harvest the crop

- Selective Harvesting: Selective harvesting involves harvesting only those parts of the crop that meet certain quality or quantity thresholds [57]. Two criteria are needed: the ability to sense the required quality factor before harvest (in-field grading) and the ability to harvest the product of interest without damaging the remaining crop. Selective harvesting presents several challenges for current robotic technology, perhaps the foremost of which is how to perform autonomous sensorimotor coordination with noisy and incomplete sensory data in the complex agricultural environment. This will likely require enhanced machine vision for recognition, segmentation, spatial localisation and tracking, as well as specialist robotic technology that is both robust and precise. Precision has traditionally been provided through stiff and easy-to-model robot mechanics. However, the increased computational resources available on robot platforms could enable some of the burden of precision to be handled through software and sensing, while allowing the robotic harvesting implements to become more passively compliant, safe and robust. Another challenge is to determine how much the robotic system should be adapted to a given crop and growing environment, and how much the growing environment should be adapted to enable better selective harvesting with robotic systems. There are interesting trade-offs to be made in this space, and which might differ from crop to crop. A related question is how to make sure the utilisation of expensive robotic hardware is maximised through the year, in particular for seasonal crops like soft fruits. Possibilities include development of adaptive technologies able to switch between tasks sharing common device capabilities, such as fruit picking and tree pruning.

Finally, there may be many additional opportunities for serendipitous parallel usage of autonomous robots alongside other field operations, for example, helping to monitor and secure farm equipment from theft and criminal damage, or protecting the sensitive habitats and species of wildlife that often coexist with agriculture.

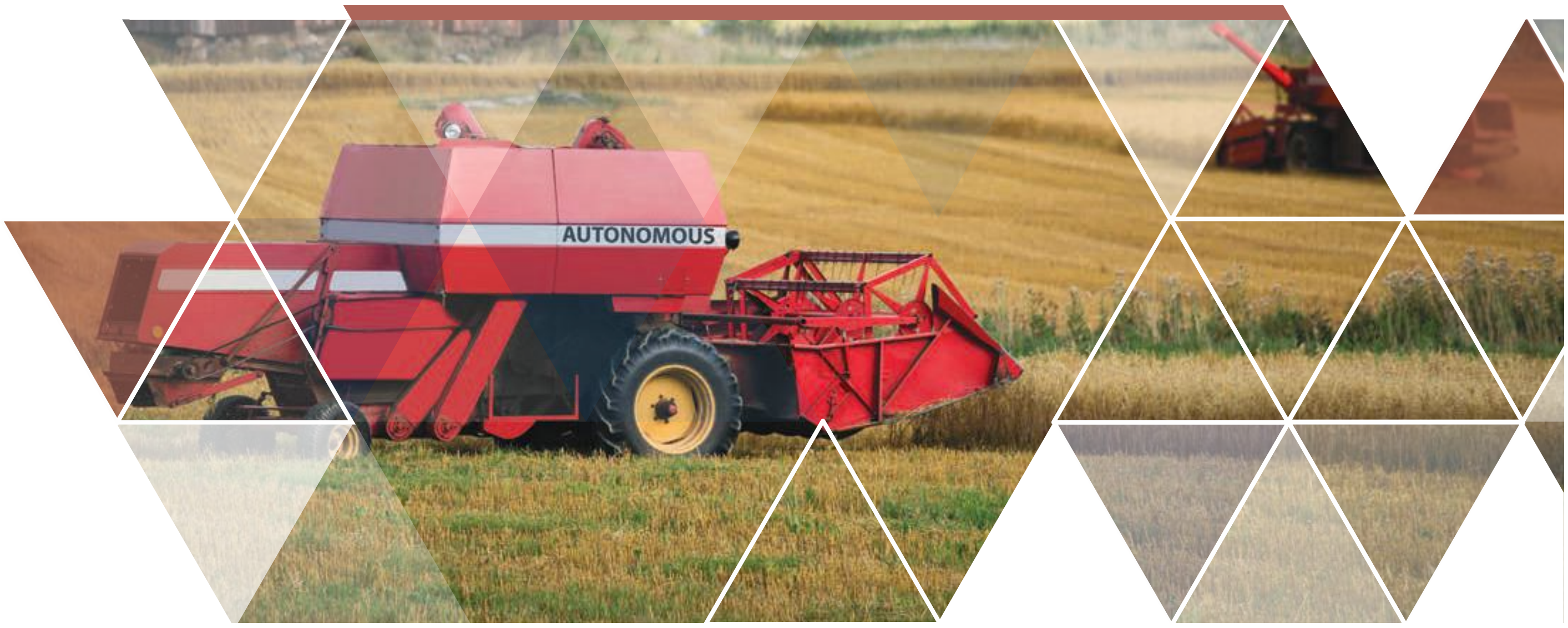

# 5. BARRIERS, CONCLUSIONS AND RECOMMENDATIONS

This white paper is about the future development of UK agriculture. RAS technologies in agriculture will become ubiquitous in the next 5 to 10 years. Robots are helping us to determine the input quantities in order to achieve desired outcomes. This white paper highlights the main trends. There are many challenges and strains in the current state of the technology for agriculture and the mechanisms for its control and governance.

However, there are particular barriers to realising the potential of RAS technology for agriculture, including the following;

1. The UK RAS community with an interest in Agri-Food RAS is small and highly dispersed. There is an urgent need to defragment and then expand the community.
2. The UK RAS community has no specific training paths or Centres for Doctoral Training to provide trained human resource capacity for RAS within Agri-Food.
3. The UK government is investing significant sums, including a new Industry Strategy Challenge Fund (ISCF) Wave 2 call (Transforming Food Production: from Farm-to-Fork), where RAS technologies have a key role. This recognises the demand, but we believe there is a danger of a mismatch where there is insufficient ongoing basic research in Agri-Food RAS to underpin onward innovation delivery for industry.
4. There is a realistic concern that RAS for Agri-Food is not realising its full potential, as the projects being commissioned currently are too few and too small-scale. RAS challenges often involve the complex integration of multiple discrete technologies (e.g. navigation, safe operation, grasping and manipulation, perception). There is a need to further develop these discrete technologies but also to deliver large scale industrial applications that resolve integration and interoperability issues.
5. The successful delivery of RAS projects within a sector domain, such as Agri-Food, requires close collaboration between the RAS community and with academic and industry practitioners. For example, the breeding of crops with novel phenotypes, such as fruits which are easy to see and pick by robots, may simplify and accelerate the application of RAS technologies. Therefore, there is an urgent need to seek new ways to create RAS and Agri-Food domain networks that can work collaboratively to address the major challenges. This is especially important for the Agri-Food domain since success in the sector always requires highly complex cross-disciplinary activity. Furthermore, within UKRI most of the Research Councils (EPSRC, BBSRC, NERC, STFC, ESRC and MRC) and Innovate-UK directly fund work in the Agri-Food domain, but as yet there is no coordinated and integrated Agri-Food research policy per se.

There is a range of technical problems that need to be addressed in a systematic and visionary manner. What are the potential solutions to these problems? The following recommendations are suggested to the government, funding agencies, industry and research centres:

- UKRI, including Research England, funding is required to train and expand human expertise for Agri-Food RAS. This may include Centres for Doctoral Training but should also provide provision for lower-level skills development through to apprentice level.
- The community needs to be defragmented. We recommends the investment in Network+ grants to stimulate and condense the community alongside the establishment of larger scale Agri-Food RAS hubs including demonstration farms. We recommend that these hubs are virtual and multi-centred, working tightly with farmers, companies and satellite universities to create the infrastructure to catalyse RAS technology. We do not believe any single UK centre currently provides the scope and capacity of expertise to deal with all the fundamental RAS challenges facing the sector. Given the global impact of Agri-Food RAS, we also recommend the UK secures international collaboration to accelerate RAS technology development.
- We recommend any new networks must comprise academic and industry Agri-Food domain expertise (crop and animal scientists, farmers, agricultural engineers), as well as representation from national RAS and government laboratories (such as the Agri-Tech Centres and Catapults), to ensure RAS solutions are compatible with industry needs. We believe many solutions to Agri-Food challenges will come from integrating RAS with more traditional technologies (agricultural engineering, crop and animal sciences, etc).
- The UK Research Councils, such as the EPSRC, STFC, ESRC, BBSRC, NERC and MRC, would benefit significantly from a coordinated Agri-Food research foresight review that integrates RAS technologies. A foresight review would help recognise the complex cross-disciplinary challenges of Agri-Food per se, but also how RAS can be integrated into a wider program. This in turn could encourage more effective responsive mode applications aligned with RAS application areas in Agri-Food.
- To deliver impact, Agri-Food RAS needs to integrate multiple technologies and resolve significant interoperability issues. We recommend that UKRI commissions a small number of large scale integration or “moon shot” projects to demonstrate routes to resolve these issues and deliver large-scale impact.
- With the new changes there is a big potential for cooperation with China, India and other countries in addressing global challenges such as sustainable food security (see, for example, the UKRI Global Challenges Research Fund). The UK has an instrumental role in this process and at the same time these new collaborations have the potential to open new avenues.
- The ongoing and large-scale government investment behind high TRL (e.g. Innovate UK led) research addressing Agri-Food sector needs is impressive. However, these investments will not succeed without investment in large-scale lower TRL research.

"Our vision is a new generation of smart, flexible, robust, compliant, interconnected robotic and autonomous systems working seamlessly alongside their human co-workers in farms and food factories. Teams of multi-modal, interoperable robotic systems will self-organise and coordinate their activities with the "human in the loop". Electric farm and factory robots with interchangeable tools, including low-tillage solutions, soft robotic grasping technologies and sensors, will support the sustainable intensification of agriculture, drive manufacturing productivity and underpin future food security."

UK-RAS NETWORK
ROBOTICS & AUTONOMOUS SYSTEMS

www.ukras.org

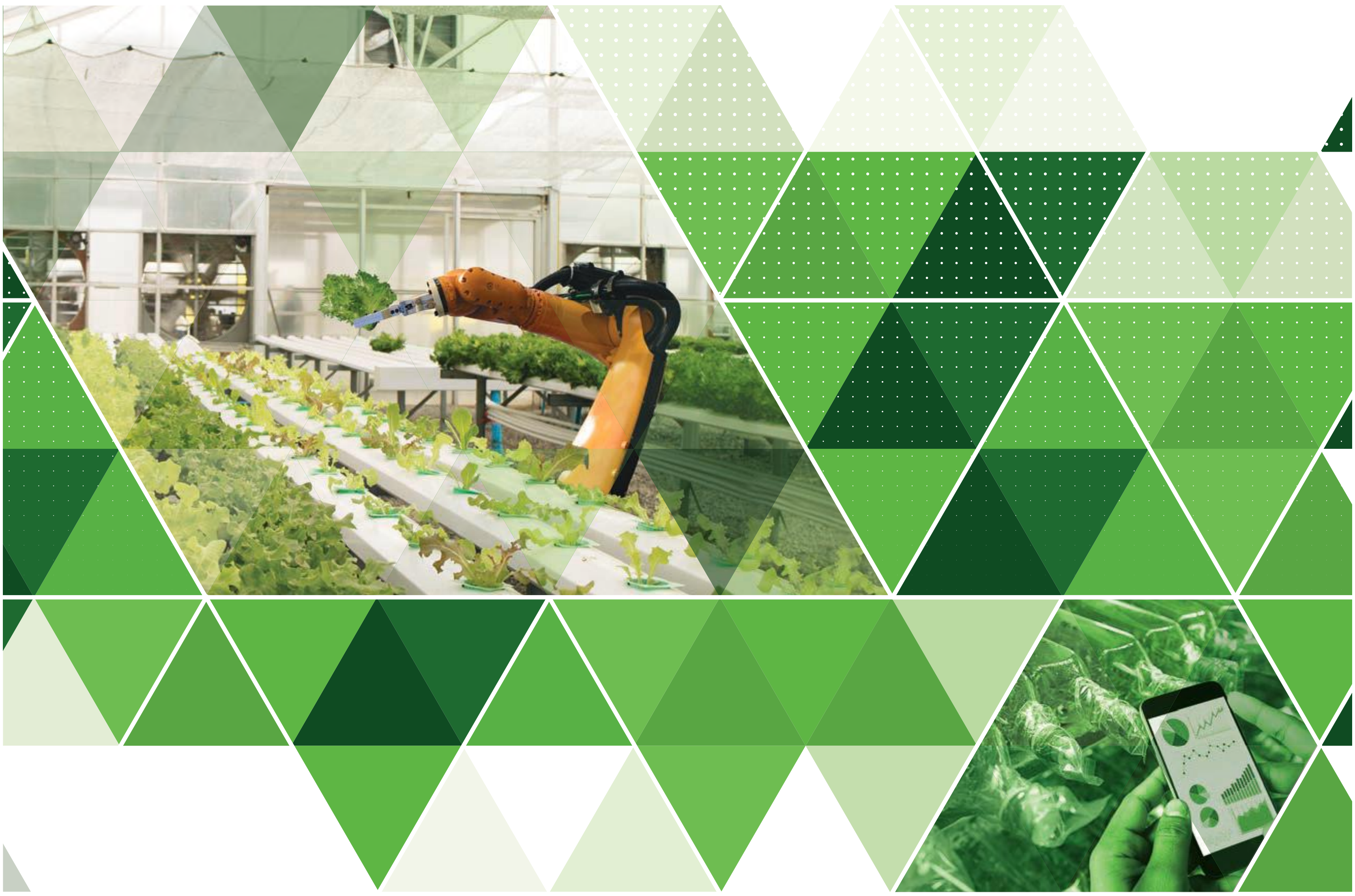